\PassOptionsToPackage{table}{xcolor}

\documentclass[twocolumn]{bytedance_seed}

\usepackage{amsmath}
\usepackage{amsfonts}
\usepackage{amssymb}

\usepackage[toc,page,header]{appendix}

\usepackage{minitoc}

\title{ S2O: Early Stopping for Sparse Attention via Online Permutation }

\author[1,*]{Yu Zhang}
\author[1,*,\dagger]{Songwei Liu}
\author[1]{Chenqian Yan}
\author[1]{Sheng Lin}
\author[2]{Beichen Ning}
\author[1,\ddagger]{Fangmin Chen}
\author[1,\ddagger]{Xing Wang}

\affiliation[1]{ByteDance}
\affiliation[2]{Xiamen University}

\contribution[*]{Equal contribution}
\contribution[\dagger]{Project leader}
\contribution[\ddagger]{Corresponding authors}

\usepackage{algorithm}
\usepackage{algpseudocode}

\definecolor{algcmt}{RGB}{46,125,122}
\newcommand{\AlgCmt}[1]{\Statex \textcolor{algcmt}{\textit{\#~#1}}}

\DeclareRobustCommand{\algsub}[1]{\ensuremath{_{\scriptscriptstyle\mathsf{#1}}}}

\algnewcommand{\algorithmicinput}{\textbf{Input:}}
\algnewcommand{\algorithmicoutput}{\textbf{Output:}}
\newcommand{\AlgInput}[1]{\Statex \algorithmicinput\ #1}
\newcommand{\AlgOutput}[1]{\Statex \algorithmicoutput\ #1}

\newcommand{\Op}[1]{\operatorname{\textsc{#1}}}
\newcommand{\Gather}{\Op{Gather}}
\newcommand{\Scatter}{\Op{Scatter}}
\newcommand{\FlashOS}{\Op{Softmax}}
\newcommand{\MaskCausal}{\Op{CausalMask}}

\DeclareMathOperator*{\Argsort}{Argsort}
\DeclareMathOperator*{\Mean}{Mean}

\newcommand{\OSupd}[4]{\FlashOS(#1,#2,#3;\,#4)}

\usepackage{booktabs}
\usepackage{multirow}
\usepackage{makecell}
\usepackage{adjustbox}

\definecolor{modelgray}{gray}{0.93}
\definecolor{oursblue}{RGB}{225,232,255}

\newcommand{\oursbg}{\cellcolor{oursblue}}
\newcommand{\oursmrow}{\textbf{Ours}}
\newcommand{\accsp}[2]{\begin{tabular}[c]{@{}c@{}}#1\\{\color{gray}\tiny #2}\end{tabular}}
\newcommand{\spcell}[2]{\makecell{#1\\{\scriptsize\textcolor{black!45}{#2}}}}
\newcommand{\spcellna}[1]{\makecell{#1\\{\scriptsize\textcolor{black!45}{--}}}}
\DeclareTextFontCommand{\textbf}{\bfseries}

\abstract{
Attention scales quadratically with sequence length, fundamentally limiting long-context inference.
Existing block-granularity sparsification can reduce latency, but coarse blocks impose an intrinsic sparsity ceiling, making further improvements difficult even with carefully engineered designs.
We present \textbf{S2O}, which performs early \textbf{s}topping for \textbf{s}parse attention via \textbf{o}nline permutation.
Inspired by virtual-to-physical address mapping in memory systems, S2O revisits and factorizes FlashAttention execution, enabling inference to load non-contiguous tokens rather than a contiguous span in the original order.
Motivated by fine-grained structures in attention heatmaps, we transform explicit permutation into an online, index-guided, discrete loading policy; with extremely lightweight preprocessing and index-remapping overhead, it concentrates importance on a small set of high-priority blocks.
Building on this importance-guided online permutation for loading, S2O further introduces an early-stopping rule: computation proceeds from high to low importance; once the current block score falls below a threshold, S2O terminates early and skips the remaining low-contribution blocks, thereby increasing effective sparsity and reducing computation under a controlled error budget.
As a result, S2O substantially raises the practical sparsity ceiling.
On \textbf{Llama-3.1-8B} under a \textbf{128K} context, S2O reduces single-operator MSE by \textbf{3.82$\times$} at matched sparsity, and reduces prefill compute density by \textbf{3.31$\times$} at matched MSE; meanwhile, it preserves end-to-end accuracy and achieves \textbf{7.51$\times$} attention and \textbf{3.81$\times$} end-to-end speedups.
}

\date{\today}
\correspondence{Songwei Liu at \email{21831068@zju.edu.cn}}

\begin{document}
\maketitle


\section{Introduction}
Large language models (LLMs) have demonstrated strong general-purpose capabilities in natural language understanding, generation, reasoning, and cross-modal tasks\cite{zhao2025surveylargelanguagemodels}.
A major driver of these gains is scaling both model size and context length.

However, as context length increases, transformer attention quickly becomes the dominant bottleneck due to its quadratic time complexity, severely limiting further progress in long-context inference.
To mitigate this issue, a growing body of work has explored sparse attention.
Among various sparsification schemes, block-sparse attention stands out as one of the most practical directions due to its simplicity and engineering feasibility:
it integrates naturally with FlashAttention~\cite{dao2023flashattention2}'s tiling pipeline and maintains high GPU utilization.
Under this paradigm, sparsity is typically applied to fixed-size compute blocks to satisfy system constraints such as aligned memory access and high parallelism.

The core challenge of block-sparse attention is deciding which attention blocks are worth computing under a limited budget.
Existing approaches can be broadly grouped into two paradigms.
\textbf{(i) Important-block selection} selects blocks under a fixed blocking scheme using predefined patterns or heuristic signals.
This includes pattern-based sparsity (e.g., A-shape patterns induced by the attention-sink phenomenon~\cite{xiao2023streamingllm}, Vertical/Slash patterns~\cite{jiang2024minference}, adaptive pattern switching~\cite{lai2025flexprefill}, and richer pattern families~\cite{li2025mminference,he2025trianglemixacceleratingprefillingdecodingtime})
and signal-driven block importance estimation (e.g., min--max sampling~\cite{tang2024quest}, anti-diagonal probing~\cite{xu2025xattention}, and similarity-based skipping~\cite{zhang2025spargeattn}).
\textbf{(ii) Permutation-based sparsification} improves efficiency by permuting token positions to cluster high-importance regions before sparse computation~\cite{xi2025sparse,yang2025sparse}.
However, in autoregressive language models, causal masking imposes strict constraints, so existing methods are typically limited to local KV permutation~\cite{wang2025sparserblocksparseattentiontoken}, which hampers global clustering.
Moreover, permutation is often applied only once prior to computation, making it difficult to iteratively refine the ordering as the attention structure changes across segments or layers.

\begin{figure*}[t]
  \centering
  \begin{minipage}[c]{0.48\textwidth}
    \centering
    \setlength{\tabcolsep}{1pt}
    \renewcommand{\arraystretch}{1.0}
    \footnotesize
    \def\imgw{0.33\linewidth}
    \begin{tabular}{@{}c@{\hspace{2pt}}c@{\hspace{2pt}}c@{}}
      \includegraphics[width=\imgw]{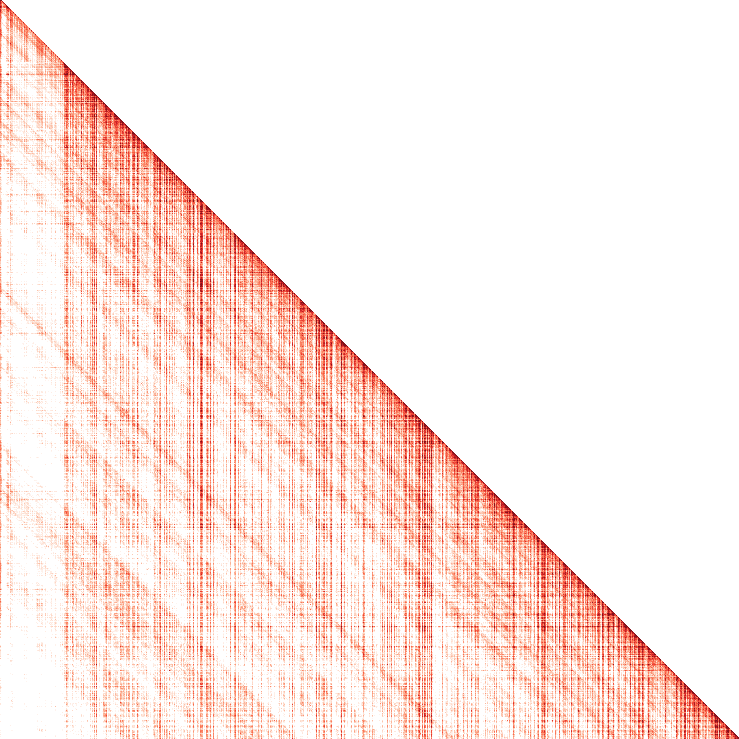}\phantomsubcaption\label{fig:attn_grid_a} &
      \includegraphics[width=\imgw]{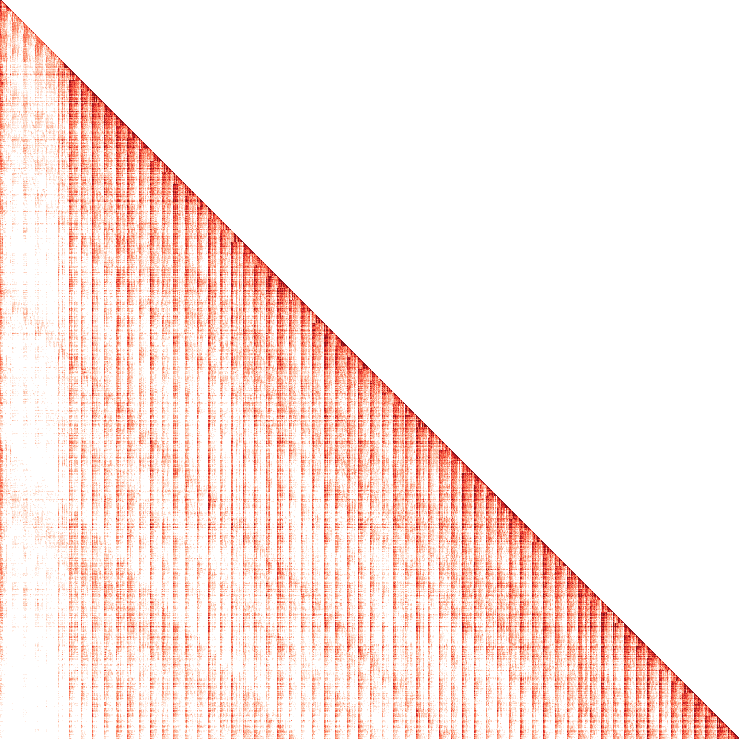}\phantomsubcaption\label{fig:attn_grid_b} &
      \includegraphics[width=\imgw]{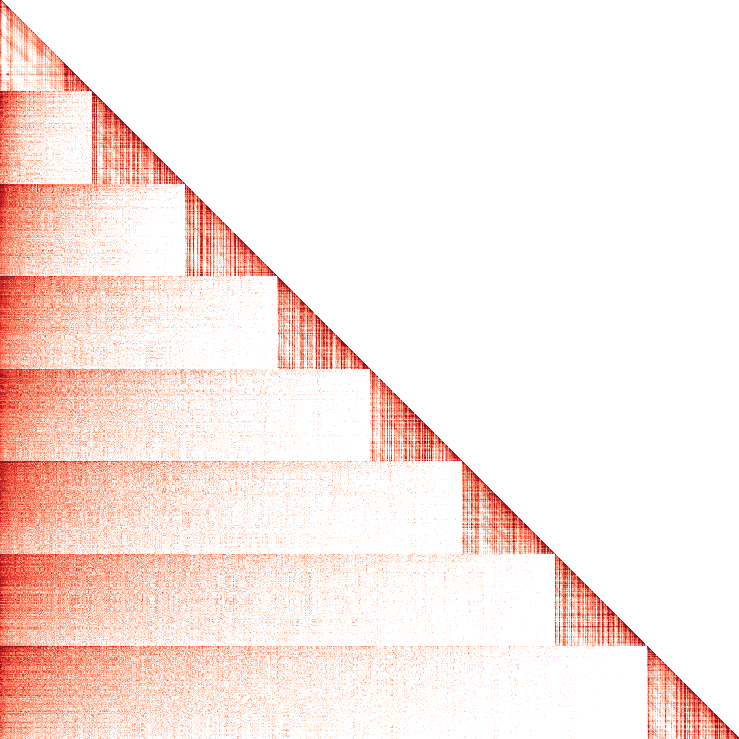}\phantomsubcaption\label{fig:attn_grid_c} \\[-2pt]
      \includegraphics[width=\imgw]{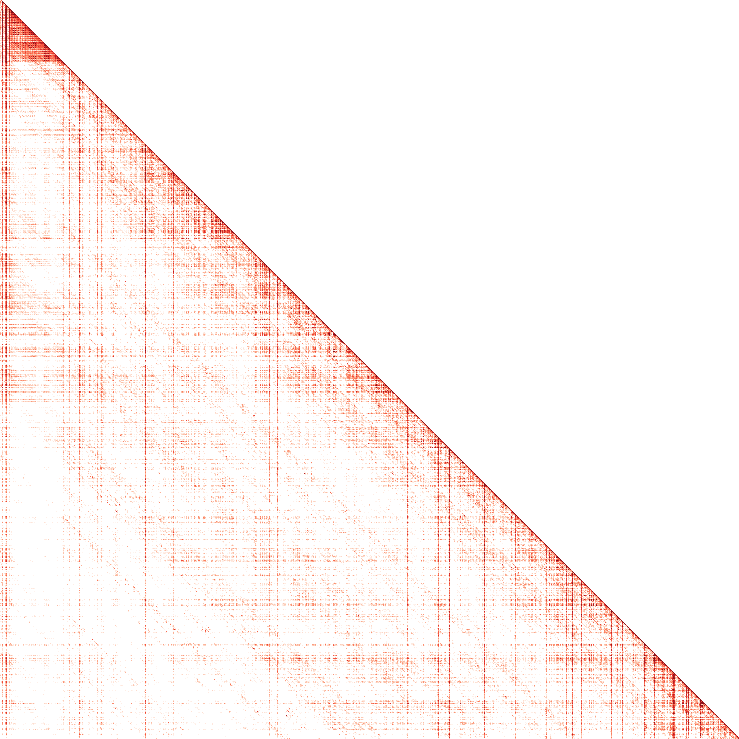} &
      \includegraphics[width=\imgw]{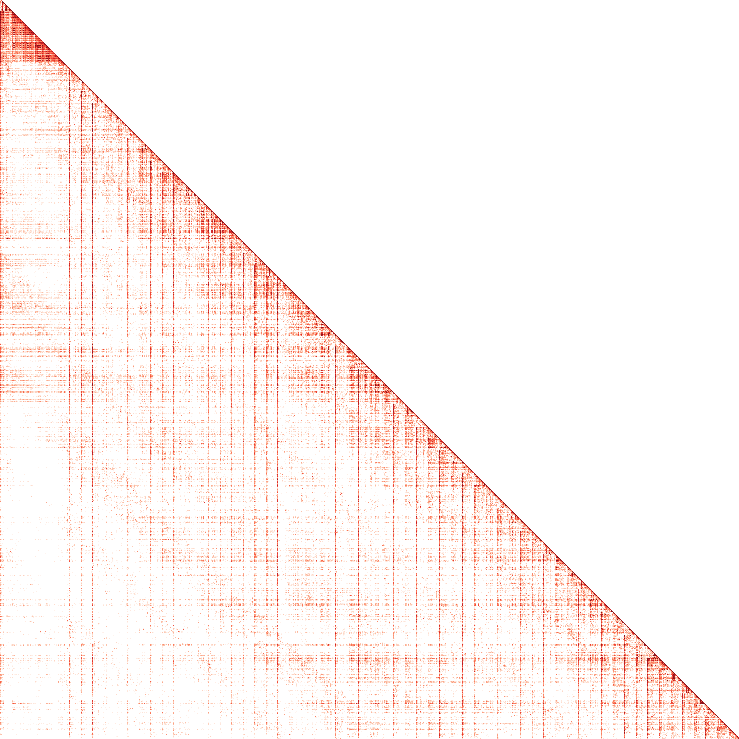} &
      \includegraphics[width=\imgw]{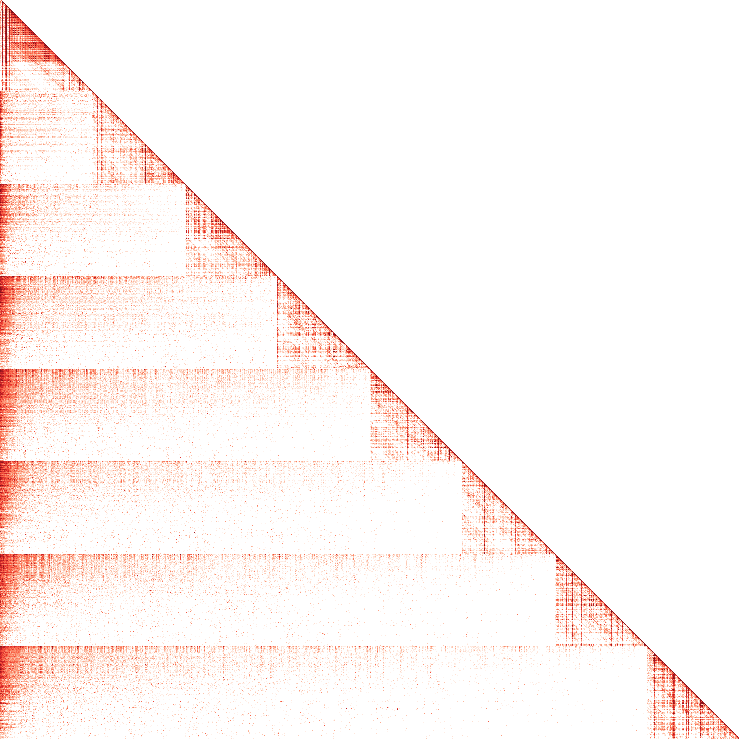} \\[-1pt]
      \textbf{(a) Original} & \textbf{(b) PBS} & \textbf{(c) Ours} \\
    \end{tabular}
    \caption[Attention heatmaps under different permutation strategies]{%
      \textbf{Attention heatmaps under different permutation strategies.}
      \textbf{(a) Original:} The heatmap exhibits abundant line-level (stripe-like) structures.
      \textbf{(b) PBS:} Following PBS's local $K/V$ permutation strategy, the heatmap still contains substantial redundancy and fails to consistently emphasize salient horizontal stripes.
      \textbf{(c) Ours:} Our global permutation scheme (Sec.~\ref{sec:seg_rank}) compacts attention mass into a progressive region from dense (upper-left) to increasingly diffuse.%
      More qualitative heatmaps are provided in Appendix~\ref{app:heatmaps}.}
    \label{fig:attn_grid}
  \end{minipage}%
  \hfill
  \begin{minipage}[c]{0.48\textwidth}
    \centering
    \includegraphics[width=\linewidth]{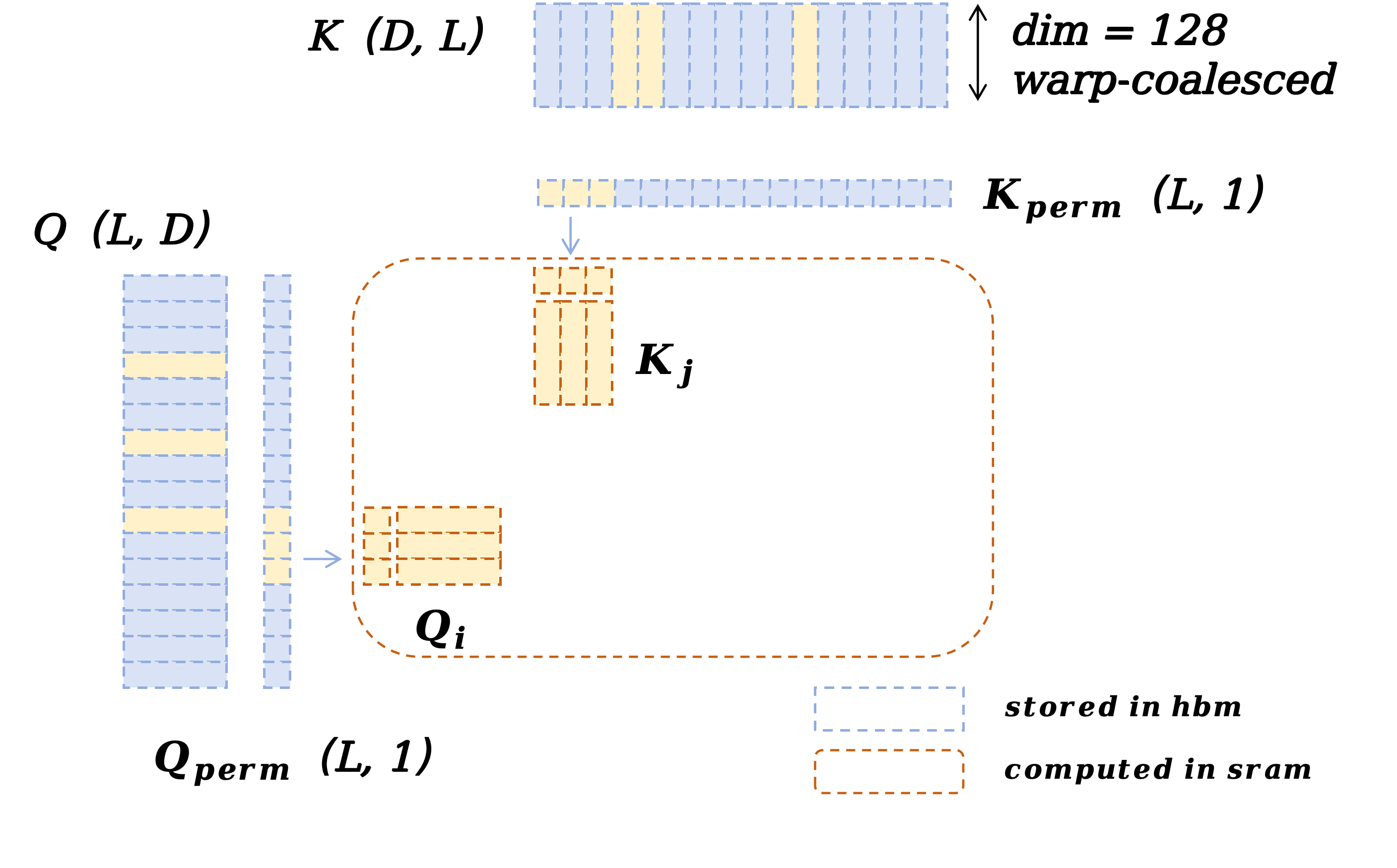}
    \caption[Coordinate-scheduled online permutation]{%
      \textbf{Coordinate-scheduled online $Q$/$K$ permutation.}
      Two lightweight index arrays $Q_{\mathrm{perm}}$ and $K_{\mathrm{perm}}$ are built without moving tensors.
      At runtime, indices directly address which tokens are loaded into on-chip SRAM for block attention.
      Since $d\!=\!128$, scattered token accesses still fully utilize a warp;
      see Appendix~\ref{sec:perm_overhead_analysis} for details.%
    }
    \label{fig:online_load}
  \end{minipage}
\end{figure*}

We make a key empirical observation: attention heatmaps in LLMs often exhibit thin, stripe-like structures rather than regular block-wise patterns (see Fig.~\ref{fig:attn_grid}).
Under coarse blocking, even accurate block selection can still waste substantial computation on low-importance positions inside selected blocks, leading to significant intra-block redundancy and a practical sparsity ceiling.
Therefore, achieving higher sparsity requires not only better block scoring, but also a mechanism that concentrates importance into a small set of blocks while avoiding the high overhead of physically permuting tokens.

\begin{figure}[t]
  \centering
  \includegraphics[width=\linewidth]{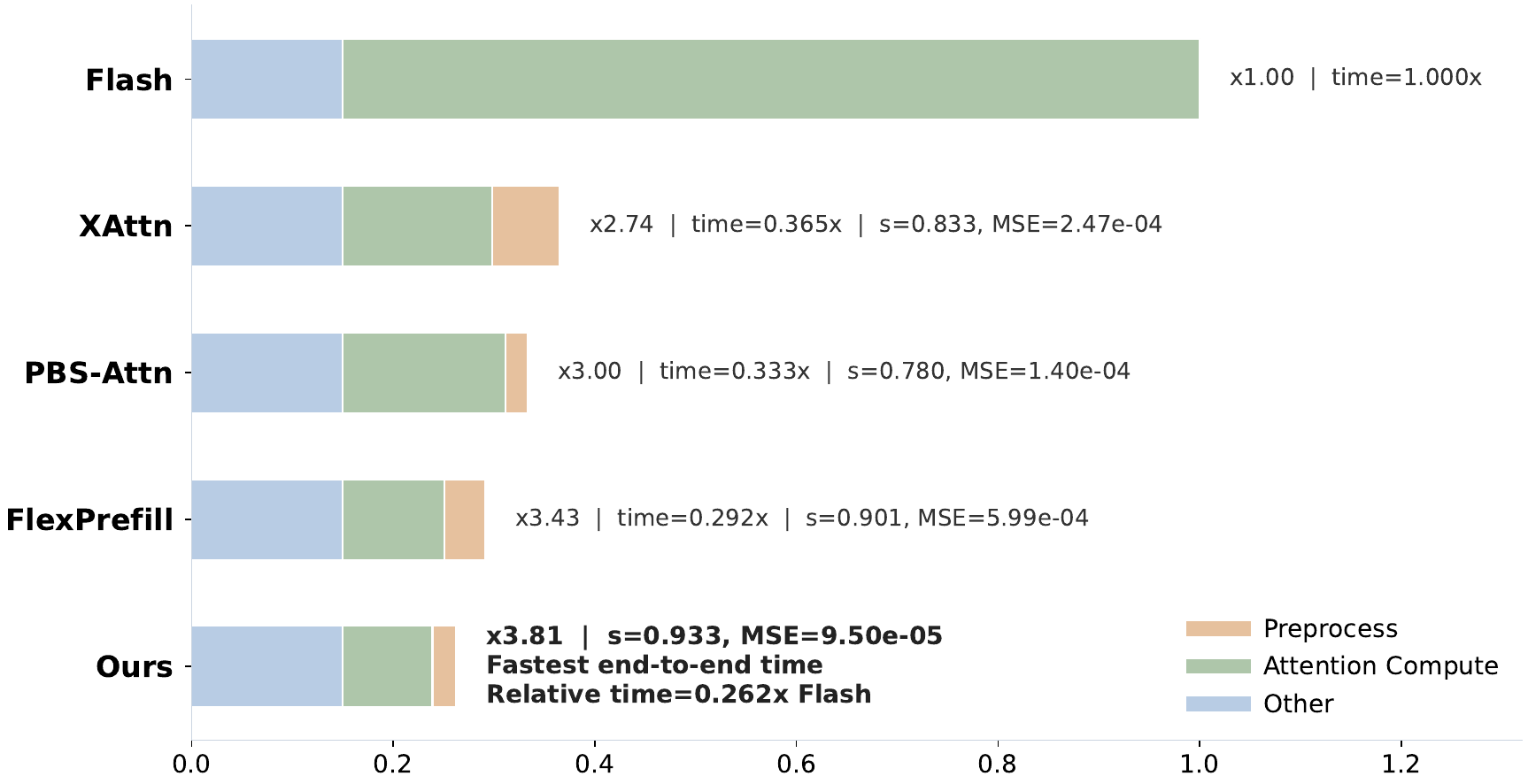}
\caption{\textbf{Speedup at 128K context length (Llama-3-8B) with sparsity--error trade-off.}
We report end-to-end latency speedup over FlashAttention and break down the major components, including sparse preprocessing time, attention compute time, and other overheads.
We also show each method's sparsity ratio and the corresponding mean squared error (MSE; lower is better), highlighting that our method achieves lower error at higher sparsity.}

  \label{fig:speed_up}
\end{figure}

To this end, we propose a FlashAttention-compatible solution with a global effect: \textbf{online permutation}.
The key idea is to preserve the physical token layout in HBM and instead introduce an extremely lightweight permutation index as a form of logical-address mapping.
Concretely, while keeping block-wise computation unchanged, we directly select and load discretely distributed query and key/value positions during the loading stage, emulating permutation benefits with negligible index-remapping overhead.
This design is based on two observations:
(i) on modern GPUs, high bandwidth utilization does not strictly require contiguous $K/V$ loads; and
(ii) the compute savings from higher sparsity can outweigh the extra memory-access overhead introduced by non-contiguous loading.
In practice, we first permute $Q$ at the segment level, aggregating highly correlated queries into a small number of contiguous $Q$ segments;
then, for each $Q$ segment, we perform multiple lightweight permutation passes over its associated $K/V$ candidates, trading a small preprocessing overhead for higher achievable sparsity.
Motivated by stripe-like attention structures~\cite{zhang-etal-2025-anchorattention}, we further introduce a stripe-granularity mean pooling signal that consolidates attention mass effectively (see Fig.~\ref{fig:attn_grid_c}), producing reliable permutation cues at low cost.

Online permutation naturally enables an \textbf{early-stopping} rule.
Because candidate $K/V$ blocks are processed in descending estimated importance under the online permutation order,
once the marginal gain from newly processed blocks becomes negligible relative to the accumulated score, we terminate early and skip the remaining low-contribution blocks.
This avoids rigidly committing to a fixed Top-$K$ subset and improves effective sparsity under a controlled error budget.

Our main contributions are as follows:
(1) We identify a practical sparsity ceiling under coarse block sparsity caused by a mismatch between stripe-like attention structures and block-wise computation, and propose \textbf{online permutation} to overcome this block-granularity limitation.
(2) We introduce a FlashAttention-compatible, index-guided loading policy with a monotone-gain \textbf{early-stopping} rule, enabling global permutation effects without physically permuting tensors.
(3) On \textbf{Llama-3-8B}\cite{grattafiori2024llama3herdmodels} under \textbf{128K} context, S2O improves the sparsity--error trade-off (\textbf{3.82$\times$} lower operator MSE or \textbf{3.31$\times$} higher sparsity), while preserving end-to-end accuracy and achieving up to \textbf{7.51$\times$} attention and \textbf{3.81$\times$} end-to-end speedups.

\section{Motivation}
\label{sec:motivation}

\subsection{How to Make the Attention Heatmap More Concentrated}
\label{sec:to_concentreated}
Across a large set of attention heatmaps, we consistently observe a pronounced dispersed pattern, as shown in Fig.~\ref{fig:attn_grid}\subref{fig:attn_grid_a}: under the standard block-sparse granularity, most blocks contain only a small number of activated weights.
From a block-level perspective, almost every block appears somewhat important, so even when block sparsification identifies relatively important blocks, substantial intra-block redundancy remains.
In other words, the bottleneck is not whether we can quickly find the most important blocks, but that a fixed block granularity has limited capacity to represent intra-block sparsity.
Therefore, rather than further debating which sparse blocks to select, it is more worthwhile to study how to reduce intra-block computation redundancy, so as to better exploit fine-grained sparsity within blocks.

\textbf{Fine-grained stripes dominate the heatmap.}
As shown in Fig.~\ref{fig:attn_grid}\subref{fig:attn_grid_a}, attention patterns often exhibit finer-grained line-level structures, which have also been reported in prior work~\cite{li2025mminference}.
The heatmap is typically shaped by interleaved vertical stripes, slash-like stripes, and horizontal stripes, among which all three can be locally salient and highly concentrated.
This phenomenon motivates three key premises for our method design:
(i) \textbf{Local vertical stripes are salient}: positions on certain vertical stripes tend to receive higher attention scores;
(ii) \textbf{Local horizontal stripes are salient}: similarly, certain horizontal stripes tend to yield higher attention scores;
(iii) \textbf{Local slash-like stripes are salient}: likewise, certain slash-like stripe patterns can also carry high attention mass.
Based on these observations, we perform mean pooling and importance estimation at the line level, rather than using fixed-size blocks as pooling units.
Moreover, compared to the PBS strategy (Fig.~\ref{fig:attn_grid}\subref{fig:attn_grid_b}) that applies local permutation over $K/V$, our approach applies a global, index-guided permutation over both $Q$ and $K/V$ (Fig.~\ref{fig:attn_grid}\subref{fig:attn_grid_c}).
This global permutation significantly concentrates attention mass toward the top-left region, yielding a more compact heatmap pattern.
Interestingly, jointly permuting $Q$, $K$, and $V$ further aligns and aggregates slash-like stripes, improving concentration.
These observations motivate a fully global, index-guided online permutation to amplify concentration.

\begin{figure*}[t]
  \centering
  \includegraphics[width=\textwidth]{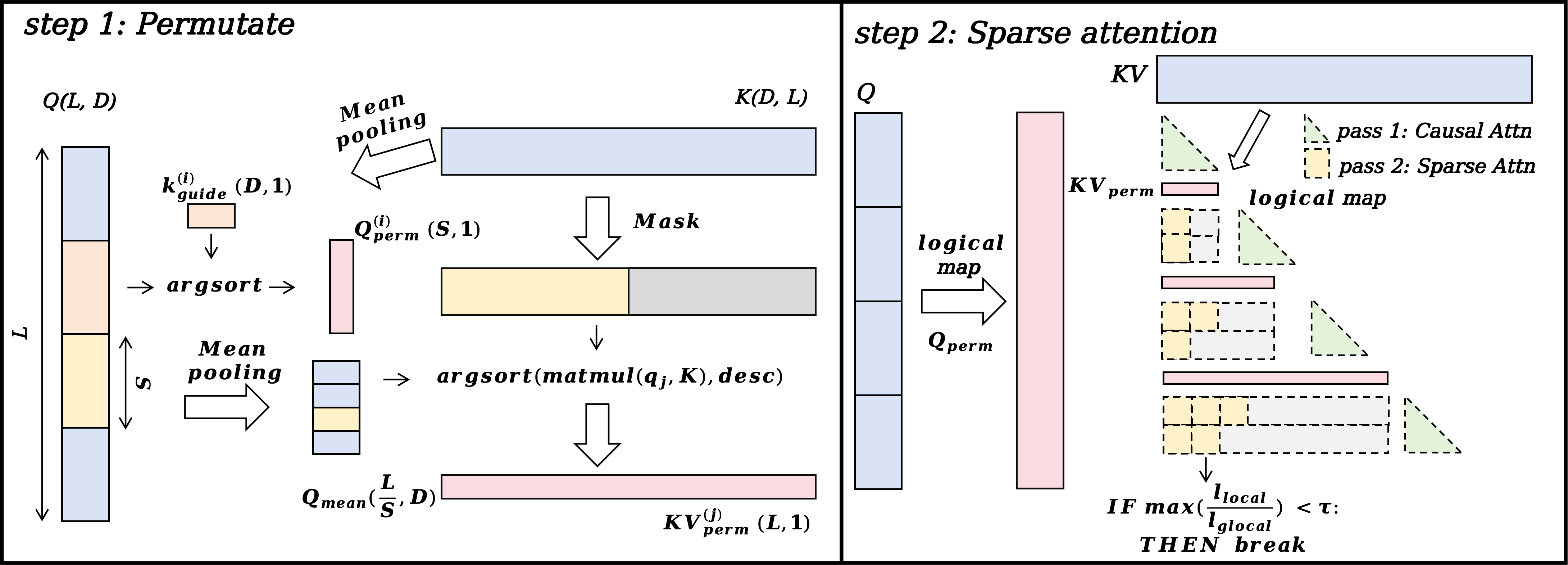}
  \caption{\textbf{S2O workflow.}
    \textbf{Step 1 (Permutate)} builds two lightweight index arrays without moving tensors:
    (i) an intra-segment query permutation index $Q_{\mathrm{perm}}$ (token-level, segment-local), and
    (ii) a prefix key/value logical map $KV_{\mathrm{perm}}$ (token-level, global indices) obtained by retrieving segment representatives.
    \textbf{Step 2 (Sparse attention)} runs attention in two passes:
    \emph{Pass-1} computes a dense intra-segment causal window to initialize online-softmax states;
    \emph{Pass-2} resumes the states and processes the historical prefix in the retrieved order with early stopping.
  }
  \label{fig:pipeline}
\end{figure*}

\subsection{How to perform online permutation}
\label{sec:motivation_online_permute}
Motivated by Section~\ref{sec:to_concentreated}, we observe that global permutation can aggregate dispersed attention mass into a more concentrated structure; to this end, we adopt \textbf{online permutation}.
We shift permutation from an \emph{offline}, pre-attention tensor reordering to a \emph{coordinate-addressed} loading procedure during computation.
As illustrated in Fig.~\ref{fig:online_load}, we first construct two lightweight index arrays, $Q_{\mathrm{perm}}$ and $K_{\mathrm{perm}}$, without moving tensors in memory.
At runtime, these indices directly determine which tiles (e.g., $Q_i$ and $K_j$) are materialized in on-chip SRAM for block-attention computation.
Compared to offline local permutation, this strategy globally front-loads high-contribution information in the loading order, enabling earlier aggregation and improving sparsification benefits.
Although token indices become non-contiguous, memory accesses within each token remain contiguous along the head-dimension, which preserves efficient GPU vectorized loads.
In practice, the added overhead is typically below \textbf{10\%} and is negligible compared to sparsity-induced speedups.
In Appendix~\ref{sec:perm_overhead_analysis}, we detail how this design integrates with FlashAttention and benchmark its runtime overhead against block-granularity sparse attention.
\section{Method}
\label{sec:method}

Motivated by Section~\ref{sec:motivation}, we leverage a key empirical pattern in long-context attention:
high-heat regions are often structured but dispersed, and permutation can rapidly concentrate them into a small set of positions.
If we compute these high-contribution positions first, the online-softmax accumulator collects most attention mass early,
enabling adaptive sparsification under a fixed approximation budget.

As shown in Figure~\ref{fig:pipeline}, our method decomposes sparse attention into two steps:
\textbf{(1) permute}---generate index arrays that permute the computation order for both $Q$ and historical $K/V$;
\textbf{(2) sparse attention}---run a FlashAttention-style kernel that gathers tokens by indices (no physical permutation),
and skips low-gain blocks via early stopping.
Crucially, we preserve the original tensor layout and memory access pattern; only lightweight index tensors are introduced.

\subsection{Step 1: Intra-Segment Lightweight Ranking}
\label{sec:seg_rank}
As $L$ grows, a direct token-level permutation over the full sequence becomes prohibitively expensive.
We therefore introduce segments to strike a better balance between benefit and overhead.
Let the sequence be partitioned into $N=L/S$ segments of length $S$.
Within each segment, we apply a cheap, index-based permutation of $Q$ to front-load queries that are more likely to induce prominent horizontal stripes;
then we use a segment representative to score and rank historical keys for that segment, producing a logical $K/V$ map and front-loading positions that are more likely to form salient vertical stripes within the segment.

\definecolor{algcomment}{RGB}{0,128,128} 

\begin{algorithm}[t]
\caption{Intra-Segment Lightweight Ranking}
\label{alg:ordering}
\footnotesize
\begin{algorithmic}[1]
\AlgInput{$Q,K\in\mathbb{R}^{Z\times H\times L\times D}$; segment length $S$}
\AlgOutput{$Q\algsub{perm}$, $KV\algsub{perm}$; }

\AlgCmt{partition the sequence into $N$ segments}
\State $N \gets L/S$; split $Q,K \rightarrow \{(Q_n,K_n)\}_{n=0}^{N-1}$

\AlgCmt{compute segment representatives}
\For{$n=0$ \textbf{to} $N-1$}
  \State $q\algsub{mean}[n] \gets \Mean(Q_n)$;\quad $k\algsub{mean}[n] \gets \Mean(K_n)$
\EndFor

\AlgCmt{(i) permute $Q$ within each segment (cheap scoring + ranking)}
\State $k\algsub{guide} \gets k\algsub{mean}[0]$
\For{$n=0$ \textbf{to} $N-1$}
  \State $s_Q[n,s] \gets \langle Q_n[s],\, k\algsub{guide}\rangle$
  \State $Q\algsub{perm}[n] \gets \Argsort(s_Q[n,:],\,\text{desc})$
\EndFor

\AlgCmt{(ii) permute historical $K/V$ per segment (causal-prefix ranking)}
\For{$n=0$ \textbf{to} $N-1$}
  \State $s_K[n,t] \gets \langle q\algsub{mean}[n],\,K[t]\rangle$
  \State $s_K[n,\,t\ge nS] \gets -\infty$ \Comment{causal prefix only}
  \State $KV\algsub{perm}[n] \gets \Argsort(s_K[n,:],\,\text{desc})$
\EndFor

\Statex \textbf{return} $Q\algsub{perm}$, $KV\algsub{perm}$
\end{algorithmic}
\end{algorithm}

\begin{algorithm}[t]
\caption{Pass-1: Dense Intra-Segment Causal Init}
\label{alg:pass1}
\footnotesize
\begin{algorithmic}[1]
\AlgInput{$Q,K,V$; segment length $S$; tile sizes $(B_M,B_N)$}
\AlgOutput{buffers $(A,Lbuf,Mbuf)$ storing $(acc,\ell,m)$ per query tile}

\AlgCmt{iterate over segments and heads}
\For{$n=0$ \textbf{to} $N-1$}
  \For{$h=0$ \textbf{to} $H-1$}
    \AlgCmt{scan queries within segment $n$ in $B_M$ tiles}
    \ForAll{\textbf{each} query tile $b$ in segment $n$}
      \State $q \gets Q(n,h,b)$
      \State $(m,\ell,acc)\gets(-\infty,0,0)$
      \AlgCmt{scan keys/values within the same segment in $B_N$ tiles}
      \ForAll{\textbf{each} key tile $t$ in segment $n$}
        \State $k \gets K(n,h,t)$;\quad $v \gets V(n,h,t)$
        \State $\mathcal{M}\gets \MaskCausal(q\text{ tile }b,\ k\text{ tile }t)$
        \State $(m,\ell,acc)\gets \OSupd{q}{k}{v}{\mathcal{M}}$
      \EndFor
      \State $A(n,h,b)\gets acc$;\quad $Lbuf(n,h,b)\gets \ell$;
      \State $Mbuf(n,h,b)\gets m$
    \EndFor
  \EndFor
\EndFor

\Statex \textbf{return} $(A,Lbuf,Mbuf)$
\end{algorithmic}
\end{algorithm}

\paragraph{(i) Segment-wise filtering of $Q$.}
We reshape $Q\in\mathbb{R}^{Z\times H\times L\times D}$ into $Q\in\mathbb{R}^{Z\times H\times N\times S\times D}$, where $N=L/S$.
For each segment, we assign every query token a coarse importance score using a lightweight guide vector.
To keep the overhead minimal, we reuse a fixed guide shared across segments (e.g., the representative key of the first segment):
\begin{equation}
  s_Q[z,h,n,s] = \big\langle Q[z,h,n,s,:],\ k_{\mathrm{guide}}[z,h,:]\big\rangle.
\end{equation}
We then compute the intra-segment permutation $Q_{\mathrm{perm}}=\mathrm{Argsort}(s_Q,\text{desc})$,
which yields segment-local offsets in $[0,S)$ and requires no tensor movement.

\paragraph{(ii) Segment-wise ranking of historical $K/V$.}
We next form a representative query for each segment by mean pooling:
$q_{\mathrm{mean}}[z,h,n,:]=\mathrm{Mean}_{s\in[0,S)}\!\big(Q[z,h,n,s,:]\big)$.
Using this representative, we score all historical keys:
\begin{equation}
  s_K[z,h,n,t] = \big\langle q_{\mathrm{mean}}[z,h,n,:],\ K[z,h,t,:]\big\rangle.
\end{equation}
To respect causality, we mask out non-prefix positions so that segment $n$ only attends to $[0,nS)$:
$s_K[z,h,n,t\ge nS]\leftarrow -\infty$.
Ranking $s_K$ in descending order yields a logical causal-prefix permutation $KV_{\mathrm{perm}}$,
i.e., an ordered list of absolute $K/V$ indices for each segment.

\subsection{Step 2: Online permutation and early skipping}
\label{sec:coord_sched}

\paragraph{Online permutation via coordinate-indexed loading.}
Rather than physically permuting tensors, we permute the computation order by supplying index arrays to the attention kernel.
We only materialize two lightweight index tensors:
$Q_{\mathrm{perm}}$, which encodes intra-segment query offsets, and $KV_{\mathrm{perm}}$, which encodes the causal-prefix $K/V$ indices.
During attention, the kernel gathers the corresponding $Q/K/V$ tiles according to these indices, executes block-wise computation in the permuted order,
and scatters the outputs back to their original positions.

\paragraph{Monotone-gain early stopping.}
In contrast to Top-$K$ or Top-CDF sampling that pre-commits to a fixed subset of blocks, we adopt an online early-stopping rule.
Following the order given by $KV_{\mathrm{perm}}$, we track the marginal attention-mass gain contributed by each processed prefix block.
Once the gain falls below a threshold $\tau$, we stop early and skip the remaining low-contribution prefix blocks, avoiding further processing of many negligible candidates without introducing an additional complex block-retrieval procedure.

\begin{algorithm}[t]
\caption{Pass-2: Coordinate-Scheduled Sparse Attention (Early Stopping)}
\label{alg:pass2}
\footnotesize
\begin{algorithmic}[1]
\AlgInput{$Q,K,V$; buffers $(A,Lbuf,Mbuf)$ (Alg.~\ref{alg:pass1});
$Q\algsub{perm},KV\algsub{perm}$ (Alg.~\ref{alg:ordering});
segment length $S$; local window $W\le S$; threshold $\tau$}
\AlgOutput{$O$}

\For{$n \gets 0$ \textbf{to} $N-1$}
  \AlgCmt{permute computation order via indices}
  \State $\tilde{Q}_n \gets \Gather(Q_n,\ Q\algsub{perm}[n])$
  
  \State $B_n \gets (A_n,Lbuf_n,Mbuf_n)$
  \State $(acc,\ell,m) \gets \Gather(B_n,\ Q\algsub{perm}[n])$

  \AlgCmt{traverse prefix tiles in $KV\algsub{perm}[n]$ with early stopping}
  \For{$T \in KV\algsub{perm}[n]$}
    \State $s \gets (m,\ell,acc)$
    \State $s' \gets \FlashOS(\tilde{Q}_n,\ K[T],\ V[T],\ s)$
    \State $(m',\ell',acc') \gets s'$
    \If{$\Delta \ell \gets \ell'-\ell;\ \Delta \ell < \tau\cdot \ell$}
      \State \textbf{break}
    \EndIf
    \State $(m,\ell,acc)\gets(m',\ell',acc')$
  \EndFor

  \State $\tilde{O}_n \gets acc/\ell$
  \State $O_n \gets \Scatter(\tilde{O}_n,\ Q\algsub{perm}[n])$
\EndFor

\Statex \textbf{return} $O$
\end{algorithmic}
\end{algorithm}

\subsection{Kernel implementation}
\label{sec:kernel_impl}
We implement Step~2 on top of a FlashAttention-style kernel.
Let $(m,\ell,acc)$ denote the per-query online-softmax states: running maximum $m$, normalization accumulator $\ell$, and output accumulator $acc$.
To avoid permutation conflicts introduced by the autoregressive causal mask, and to ensure numerical stability, the kernel runs in two passes (Figure~\ref{fig:pipeline}).

\paragraph{Pass~1: Dense Intra-Segment Causal Init.}
For each segment, we first compute a small dense causal window within the segment (Algorithm~\ref{alg:pass1}).
This is because, under the autoregressive causal mask, permutation may bring originally masked-out positions forward, making dense-block computation and scheduling more complicated;
the intra-segment dense window initializes stable online-softmax states for each query token and caches them in lightweight buffers $(A,Lbuf,Mbuf)$.

\paragraph{Pass~2:  Coordinate-Scheduled Sparse Attention (Early Stopping).}
We then permute queries inside the segment according to $Q_{\mathrm{perm}}$, gather the corresponding $Q$ tokens, and resume states from $(A,Lbuf,Mbuf)$.
Next, we traverse the causal prefix keys in the order specified by $KV_{\mathrm{perm}}$ using tiles of size $B_N$,
updating $(m,\ell,acc)$ with the standard online-softmax recurrence.
After each tile, we estimate the marginal gain in the normalization mass $\ell$; if $\Delta \ell$ falls below a fraction of the accumulated $\ell$, we terminate early.
Finally, we normalize $o=acc/\ell$ and scatter the outputs back to their original token positions.

\begin{figure*}[t]
  \centering
  \setlength{\tabcolsep}{6pt}
  \begin{tabular}{c c c}
    \begin{subfigure}[t]{0.32\textwidth}
      \centering
      \includegraphics[width=\linewidth]{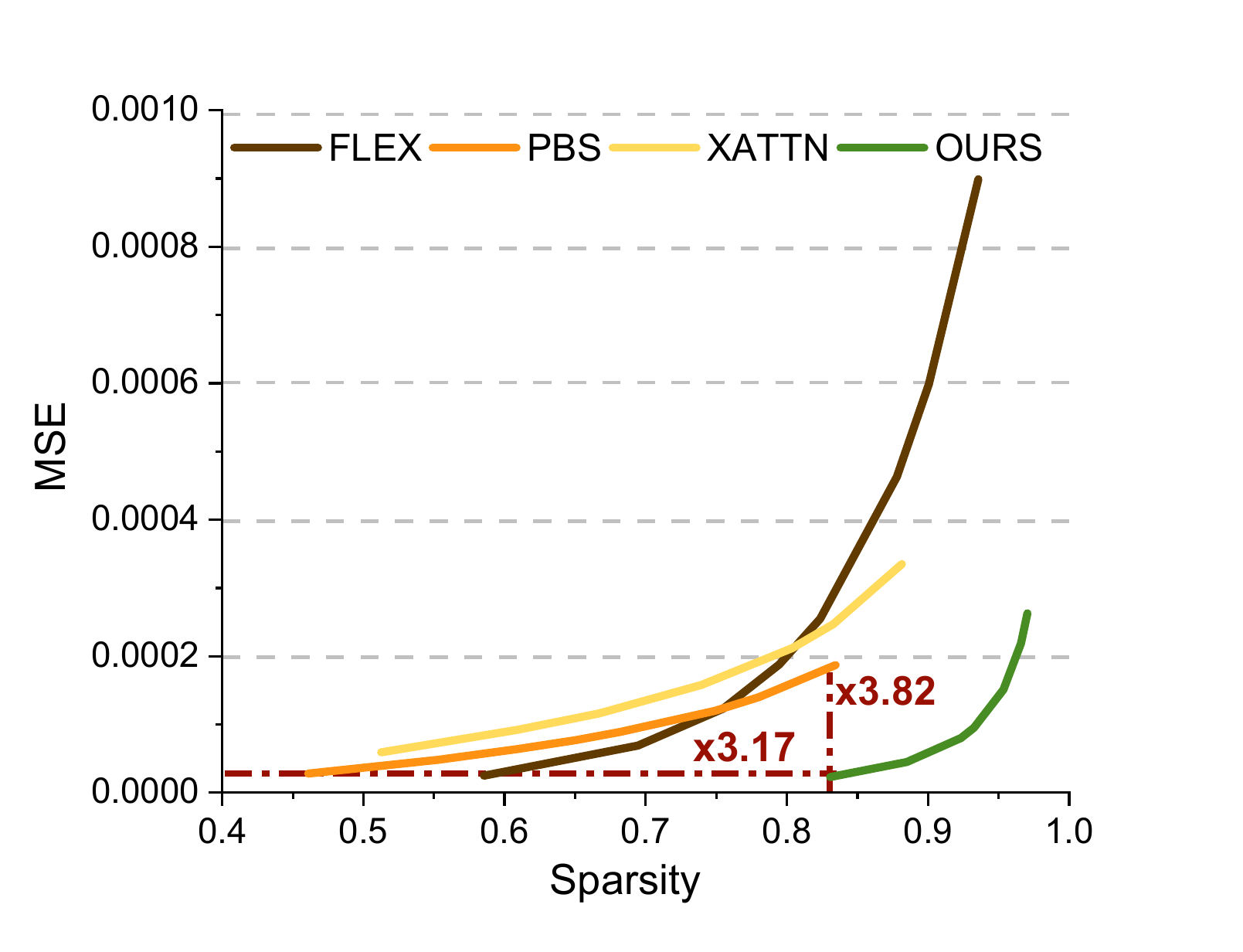}
      \caption{Mse vs. Sparsity.}
      \label{fig:loss_sparsity}
    \end{subfigure}
    &
    \begin{subfigure}[t]{0.32\textwidth}
      \centering
      \includegraphics[width=\linewidth]{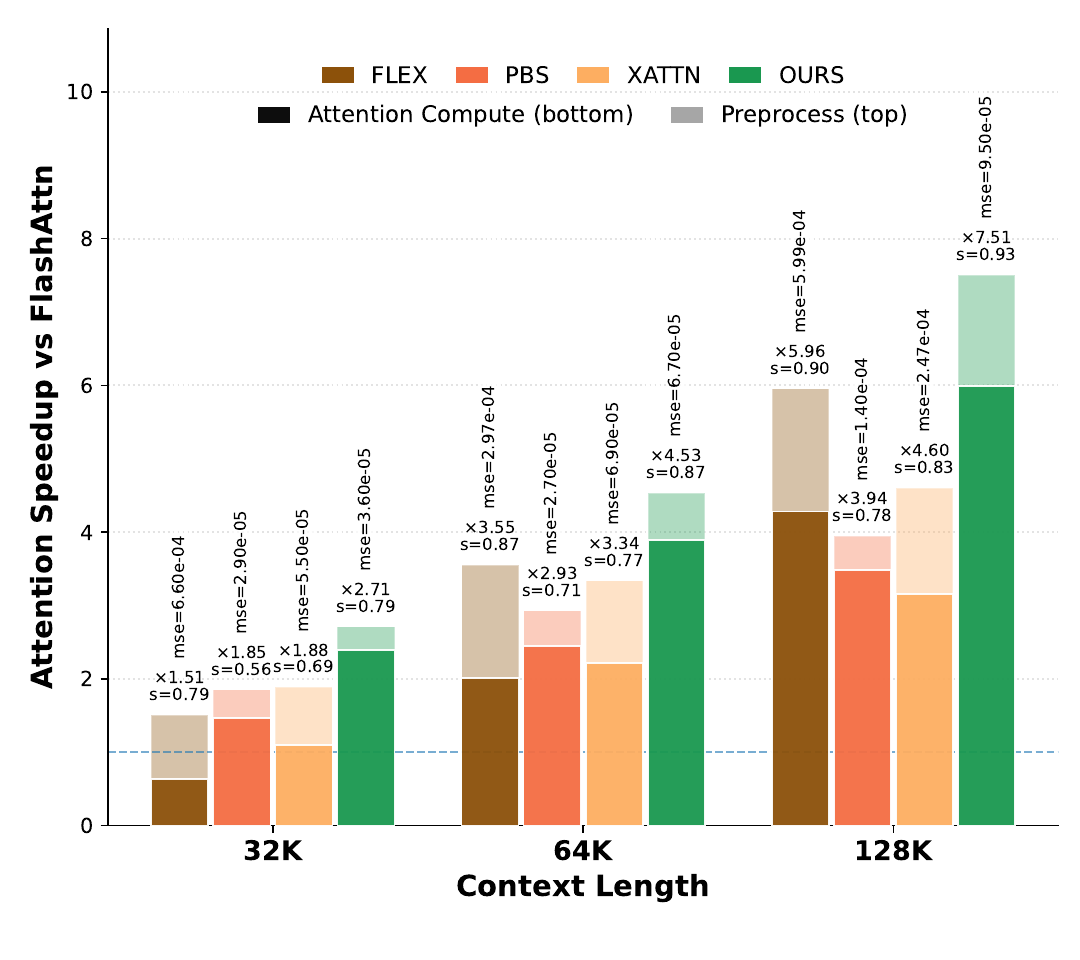}
      \caption{Attention Speedup.}
      \label{fig:retrieval_overhead}
    \end{subfigure}
    &
    \begin{subfigure}[t]{0.32\textwidth}
      \centering
      \includegraphics[width=\linewidth]{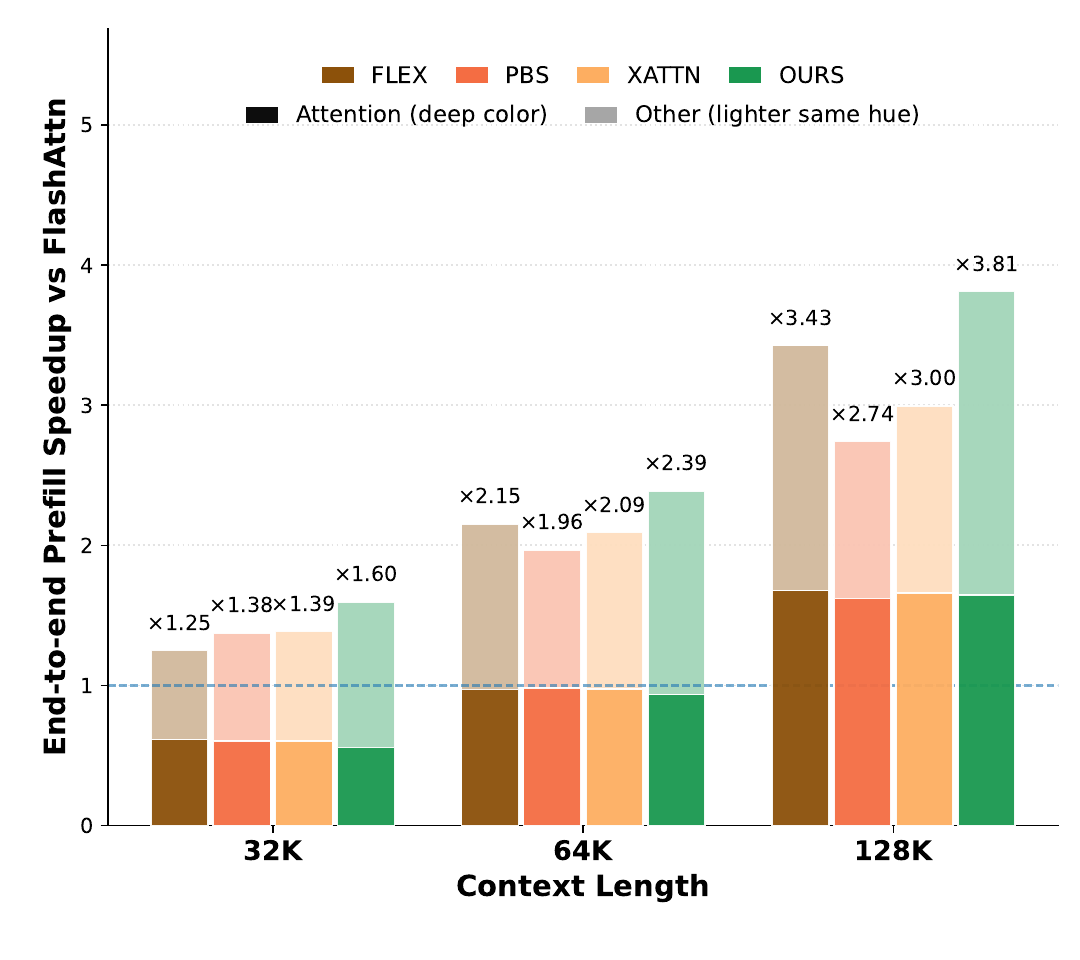}
      \caption{E2e Speedup.}
      \label{fig:e2e_breakdown}
    \end{subfigure}
  \end{tabular}

  \caption{Evaluation overview on \textbf{Llama-3.1-8B}. (a) Operator-level accuracy under matched sparsity. (b) Attention speedup: dark bars denote attention time and light bars denote preprocessing time; the corresponding sparsity(s) and MSE are also reported. (c) End-to-end prefill latency breakdown across context lengths: dark bars denote total attention-related time (including preprocessing) and light bars denote all remaining overheads.}
  \label{fig:eval_overview}
\end{figure*}

\newcommand{\best}[1]{\textbf{#1}}
\newcommand{\second}[1]{\underline{#1}}

\definecolor{oursblue}{RGB}{230,235,255}

\definecolor{modelgray}{gray}{0.93}
\definecolor{oursblue}{RGB}{225,232,255}

\providecommand{\oursbg}{\cellcolor{oursblue}}
\providecommand{\oursmrow}{\textbf{Ours}}

\providecommand{\accsp}[2]{\begin{tabular}[c]{@{}c@{}}#1\\{\color{gray}\tiny #2}\end{tabular}}

\begin{table*}[t]
  \centering
  \scriptsize
  \renewcommand{\arraystretch}{1.05}
  \begin{adjustbox}{width=\textwidth}
  \begin{tabular}{l *{10}{c}}
    \toprule
    Methods &
    En.Sum & En.QA & En.MC & En.Dia & Zh.QA &
    Code.Debug & Math.Find &
    Retr.PassKey & Retr.Number &
    Avg. \\
    \midrule

    \rowcolor{modelgray}
    \multicolumn{11}{c}{\textbf{Llama-3.1-8B}} \\
    \midrule
    Full &
    \accsp{0.167}{0.0000} & \accsp{0.187}{0.0000} & \accsp{0.271}{0.0000} & \accsp{0.135}{0.0000} & \accsp{0.132}{0.0000} &
    \accsp{0.241}{0.0000} & \accsp{0.331}{0.0000} &
    \accsp{0.993}{0.0000} & \accsp{0.993}{0.0000} &
    \accsp{0.383}{0.0000} \\
    FlexPrefill &
    \accsp{0.167}{0.8591} & \accsp{0.178}{0.8523} & \accsp{0.245}{0.8534} & \accsp{0.125}{0.8416} & \accsp{0.113}{0.8480} &
    \accsp{0.218}{0.8376} & \accsp{0.328}{0.8682} &
    \accsp{\textbf{1.000}}{0.8760} & \accsp{0.988}{0.8510} &
    \accsp{0.374}{0.8541} \\
    XAttn &
    \accsp{0.180}{0.7106} & \accsp{0.165}{0.7132} & \accsp{0.223}{0.7128} & \accsp{\textbf{0.135}}{0.7247} & \accsp{0.080}{0.6831} &
    \accsp{0.234}{0.7303} & \accsp{0.329}{0.7075} &
    \accsp{0.998}{0.6892} & \accsp{0.995}{0.6856} &
    \accsp{0.371}{0.7063} \\
    PBS &
    \accsp{\textbf{0.179}}{0.7539} & \accsp{\textbf{0.192}}{0.7578} & \accsp{0.245}{0.7570} & \accsp{0.115}{0.7709} & \accsp{\textbf{0.127}}{0.7292} &
    \accsp{0.228}{0.7766} & \accsp{0.329}{0.7513} &
    \accsp{0.831}{0.7415} & \accsp{0.992}{0.7376} &
    \accsp{0.360}{0.7529} \\

    \rowcolor{oursblue}
    \oursmrow &
    \accsp{0.170}{\textbf{0.9178}} & \accsp{0.166}{\textbf{0.9220}} & \accsp{\textbf{0.271}}{\textbf{0.9225}} & \accsp{0.120}{\textbf{0.9150}} & \accsp{0.110}{\textbf{0.9106}} &
    \accsp{\textbf{0.244}}{\textbf{0.9096}} & \accsp{\textbf{0.343}}{\textbf{0.9154}} &
    \accsp{0.986}{\textbf{0.9390}} & \accsp{\textbf{0.998}}{\textbf{0.9382}} &
    \accsp{\textbf{0.379}}{\textbf{0.9211}} \\
    \midrule

    \rowcolor{modelgray}
    \multicolumn{11}{c}{\textbf{Qwen3-8B}} \\
    \midrule
    Full &
    \accsp{0.180}{0.0000} & \accsp{0.022}{0.0000} & \accsp{0.489}{0.0000} & \accsp{0.275}{0.0000} & \accsp{0.038}{0.0000} &
    \accsp{0.185}{0.0000} & \accsp{0.317}{0.0000} &
    \accsp{1.000}{0.0000} & \accsp{0.969}{0.0000} &
    \accsp{0.386}{0.0000} \\
    FlexPrefill &
    \accsp{0.149}{0.8250} & \accsp{0.019}{0.8266} & \accsp{0.450}{0.8494} & \accsp{0.160}{0.8329} & \accsp{\textbf{0.039}}{0.8328} &
    \accsp{0.170}{0.8581} & \accsp{\textbf{0.340}}{0.8164} &
    \accsp{0.998}{0.8181} & \accsp{0.990}{0.8237} &
    \accsp{0.368}{0.8314} \\
    XAttn &
    \accsp{0.164}{0.6631} & \accsp{0.019}{0.6701} & \accsp{0.415}{0.6697} & \accsp{0.165}{0.6914} & \accsp{0.037}{0.6669} &
    \accsp{0.200}{0.7147} & \accsp{0.306}{0.6189} &
    \accsp{1.000}{0.6229} & \accsp{0.810}{0.6212} &
    \accsp{0.346}{0.6599} \\

    PBS &
    \accsp{\textbf{0.180}}{0.7730} & \accsp{0.019}{0.7468} & \accsp{\textbf{0.476}}{0.7473} & \accsp{\textbf{0.305}}{0.7649} & \accsp{0.038}{0.7424} &
    \accsp{0.172}{0.7806} & \accsp{0.337}{0.7098} &
    \accsp{1.000}{0.7175} & \accsp{\textbf{1.000}}{0.7209} &
    \accsp{\textbf{0.392}}{0.7504} \\
    \rowcolor{oursblue}
    \oursmrow &
    \accsp{0.175}{\textbf{0.9085}} & \accsp{\textbf{0.020}}{\textbf{0.9089}} & \accsp{0.445}{\textbf{0.8980}} & \accsp{0.280}{\textbf{0.8669}} & \accsp{0.038}{\textbf{0.8663}} &
    \accsp{\textbf{0.200}}{\textbf{0.8745}} & \accsp{0.323}{\textbf{0.8736}} &
    \accsp{\textbf{1.000}}{\textbf{0.8554}} & \accsp{0.997}{\textbf{0.9033}} &
    \accsp{0.386}{\textbf{0.8839}} \\

    \bottomrule
  \end{tabular}
  \end{adjustbox}
  \caption{\textbf{InfiniteBench results.}
  Each cell shows Accuracy (top) and Prefill-stage sparsity ratio (bottom) under identical decoding settings.
\textbf{Bold} indicates the sparsity strategy with the highest score for each task.} 
  \label{tab:infinitebench_acc_spars}
\end{table*}

\section{Experiments}
\label{sec:exp}

\subsection{Experimental Setup}
\label{sec:exp_setup}
\paragraph{Models.}
We evaluate on two widely used open-source LLMs, \textbf{Qwen3-8B}~\cite{yang2025qwen3technicalreport} and \textbf{Llama-3.1-8B}~\cite{grattafiori2024llama3herdmodels}. They are representative of mainstream deployments and capture typical attention-heatmap behaviors observed in modern LLMs.

\paragraph{Baselines.}
We compare our method against:
(i) \textbf{Full} (implemented with \textbf{FlashAttention2}~\cite{dao2023flashattention2}), used as the dense-attention reference;
(ii) \textbf{FlexPrefill}~\cite{lai2025flexprefill}: it dynamically determines a per-head sparse pattern by comparing estimated vs.\ true score distributions, representing pattern-based block selection; 
(iii) \textbf{XAttn}~\cite{xu2025xattention}: a training-free block-sparse method that scores blocks via an anti-diagonal probing pattern, representing a fine-grained block selection strategy;
and
(iv) \textbf{PBS}~\cite{wang2025sparserblocksparseattentiontoken}: it leverages attention’s permutation property and performs segment-wise token permutation to concentrate high-mass regions, representing a local permutation strategy.

\paragraph{Implementation details.}
\label{iml_details}
All experiments are conducted in BF16 on NVIDIA GPUs.
We implement our method in Triton following a FlashAttention-style kernel structure.
Unless otherwise specified, we use segment length $S{=}2048$, early-stop threshold $\tau{=}0.005$, and fix the attention block size to $(B_M,B_N)=(128,128)$ for fair comparison across methods.
Latency is measured using CUDA events, and we report the mean over multiple post-warmup iterations. MSE/MAE are computed as the mean error averaged across all attention heads.
All baselines use the default configurations reported in their papers (see Appendix~\ref{sec:append} for details).

\subsection{Main Results}
\label{sec:exp_main}

\paragraph{Single-operator and end-to-end performance.}
\noindent We tune each method's hyperparameters to control its attained sparsity (i.e., compute density), and compare the resulting approximation error under approximately matched sparsity.
To quantify operator-level approximation quality, we introduce mean squared error (MSE) as a direct metric for sparse-attention error: a lower MSE indicates that the attention output is closer to the full-attention reference.
As shown in Fig.~\ref{fig:loss_sparsity}, at matched sparsity, our method consistently achieves substantially lower MSE than all baselines.
On \textbf{Llama-3.1-8B} with a \textbf{128K} context, compared to the prior state-of-the-art strategy, our method reduces MSE by up to \textbf{3.82$\times$} at matched sparsity, and reduces the required compute density by up to \textbf{3.17$\times$} at matched MSE, demonstrating superior single-operator quality.

\noindent Beyond the sparse-attention operator itself, sparse preprocessing (e.g., retrieval/indexing) may introduce non-negligible overhead in practical systems.
We therefore profile the end-to-end prefill latency under the default configuration across different sequence lengths, and decompose the total time into
(i) attention-operator time, (ii) sparse-preprocessing time, and (iii) end-to-end prefill time.
As shown in Table~\ref{tab:retrieval_overhead_seglen_seqlen}, under long-context settings, our preprocessing overhead accounts for only a small fraction of the total time and is dominated by a single lightweight ordering step.
Moreover, Fig.~\ref{fig:retrieval_overhead},  Fig.~\ref{fig:e2e_breakdown} and Fig.~\ref{fig:speed_up} show consistent speedups across a wide range of context lengths, indicating that operator-level improvements (as measured by MSE) effectively translate into system-level gains under realistic workloads.
In particular, under a \textbf{128K} context on \textbf{Llama-3.1-8B}, our method achieves \textbf{7.53$\times$} operator-level attention speedup and \textbf{3.81$\times$} end-to-end speedup.

\paragraph{Benchmark results.}
We focus on long-context benchmarks and evaluate accuracy on
\textsc{RULER}, \textsc{LongBench v2}, and \textsc{InfiniteBench}.
The main results are summarized in Tab.~\ref{tab:infinitebench_acc_spars} (\textsc{InfiniteBench}),
Tab.~\ref{tab:ruler_acc_sparsity} (\textsc{RULER}~\cite{hsieh2024ruler}),
and Tab.~\ref{tab:longbenchv2} (\textsc{LongBench v2}~\cite{bai2024longbench2}).
Due to space constraints, additional results are deferred to Appendix~\ref{sec:append_hyper}.
In particular, Tab.~\ref{tab:ruler_acc_sparsity} shows that on \textsc{RULER}, our method maintains accuracy across a wide range of context lengths while achieving higher sparsity.
Overall, our method attains accuracy comparable to \textsc{Full-Attn} across all benchmarks,
while reducing the prefill compute density by \textbf{2--3$\times$} compared to the original strategy,
thereby providing a higher-ceiling path for sparse-attention design.

\subsection{Ablation Study}
\label{sec:exp_ablation}

\paragraph{Hyperparameter ablations.}
We conduct hyperparameter ablations on \textbf{Qwen3-8B} under a \textbf{128K} context by varying the segment length $S$ and the early-stop threshold $\tau$, and report their impacts on sparsity as well as mean squared error (MSE) and mean absolute error (MAE).
Overall, $S$ has only a minor effect on approximation error, whereas $\tau$ is the dominant factor governing the speed--accuracy trade-off.
Based on these results, we use the recommended $S$ in the table to balance retrieval overhead, and adjust $\tau$ to trade accuracy for sparsity gains.

\begin{table}[t]
  \centering
  \scriptsize
  \renewcommand{\arraystretch}{1.05}
  \resizebox{\columnwidth}{!}{%
  \begin{tabular}{ccccc}
    \toprule
    \multicolumn{5}{c}{\textbf{Hyperparameter Ablations (128K)}} \\
    \midrule
    $S$ & $\tau$ & Sparsity$\uparrow$ & avg.\ MAE$\downarrow$ & avg.\ MSE$\downarrow$ \\
    \midrule
    2048 & 0.001 & 0.721 & 0.00536 & 0.00058 \\
    2048 & 0.002 & 0.798 & 0.00873 & 0.00131 \\
    2048 & 0.004 & 0.861 & 0.01353 & 0.00265 \\
    2048 & 0.005 & 0.877 & 0.01535 & 0.00323 \\
    2048 & 0.010 & 0.918 & 0.02163 & 0.00560 \\
    2048 & 0.020 & 0.944 & 0.02849 & 0.00876 \\
    \midrule
    1024 & 0.005 & 0.887 & 0.01519 & 0.00322 \\
    2048 & 0.005 & 0.877 & 0.01535 & 0.00323 \\
    4096 & 0.005 & 0.862 & 0.01520 & 0.00312 \\
    \bottomrule
  \end{tabular}%
  }
  \caption{\textbf{Hyperparameter ablations.}
  We report mean sparsity and the average approximation error (MAE/MSE, averaged over heads) on Qwen3-8B at 128K.
  Top block fixes $S{=}2048$ and varies $\tau$; bottom block fixes $\tau{=}0.005$ and varies $S$.}
  \label{tab:ablation_hparams_128k_l1l2}
\end{table}

\begin{table}[t]
  \centering
  \scriptsize
  \renewcommand{\arraystretch}{1.08}
  \resizebox{\columnwidth}{!}{
  \begin{tabular}{lccc}
    \toprule
    Method & Sparsity $\uparrow$ & MAE $\downarrow$ & MSE $\downarrow$ \\
    \midrule
    \textsc{local window only}                          & 0.983 & 0.01790 & 1.8395$\times10^{-3}$ \\
    \textsc{Ours (w/o $Q_{\mathrm{perm}}$)}        & 0.796 & 0.00133 & 1.86$\times10^{-5}$ \\
    \rowcolor{oursblue}
    \textbf{Ours (w. $Q_{\mathrm{perm}}$)}        & 0.852 & 0.00129 & 1.87$\times10^{-5}$ \\
    \bottomrule
  \end{tabular}}
  \caption{\textbf{Component ablation.}
  We report mean sparsity and average approximation error (MAE/MSE) on Qwen3-8B at 128K, and additionally present one representative head.}
  \label{tab:ablation_comp_128k}
\end{table}

\paragraph{Component ablation.}
To quantify the contribution of key design choices, we ablate two core components:
\textbf{(i) the local dense window} for stabilizing attention near segment boundaries, and
\textbf{(ii) query-side intra-segment permutation} $Q_{\mathrm{perm}}$.
We compare the full method with two variants (Table~\ref{tab:ablation_comp_128k}):
\emph{w/o} $Q_{\mathrm{perm}}$ (setting $Q_{\mathrm{perm}}$ to the identity order within each segment while keeping the same historical-prefix $K/V$ ordering and early-stopping rule), and \emph{local-window only} (retaining only dense local-window attention).
On a subset of heads, enabling $Q_{\mathrm{perm}}$ consistently improves the speed--quality trade-off by front-loading queries that are more likely to induce horizontal stripes, allowing earlier mass accumulation and more effective early stopping at comparable error.
In contrast, using the local dense window alone incurs large approximation error; its main role is to avoid special-case handling of masked positions near segment boundaries. For weaker stripe patterns, we also provide a simplified variant that performs no $Q$ reordering and fuses pass-1 and pass-2 into a single computation (see Alg.~\ref{alg:fused_pass12_noqreorder}).
Since this variant has nearly identical runtime to the full implementation, we ultimately adopt a unified $Q_{\mathrm{perm}}$ permutation strategy across all heads.

\section{Related Work}
\label{sec:related_work}

\paragraph{Sparse attention.}
FlashAttention~\cite{dao2023flashattention2}, inspired by online-softmax~\cite{milakov2018onlinenormalizercalculationsoftmax}, reduces HBM traffic through blockwise loading and incremental accumulation, providing a practical foundation for block-sparse attention.
Train-free sparse attention restricts computation to structured regions without retraining, including local-window/global-token schemes~\cite{xiao2023streamingllm,beltagy2020longformerlongdocumenttransformer},
pattern-based block sparsity~\cite{jiang2024minference,li2025mminference,he2025trianglemixacceleratingprefillingdecodingtime,lai2025flexprefill},
and improved block selection/sampling strategies~\cite{tang2024quest,xu2025xattention,zhang2025spargeattn,gu2025bladeblocksparseattentionmeets}.
Moreover, prior work shows that reordering can concentrate attention mass and improve sparse access patterns~\cite{xi2025sparse,yang2025sparse,wang2025sparserblocksparseattentiontoken}.

\paragraph{Efficient Transformers.}
\textbf{Quantization} reduces memory footprint and bandwidth via low-bit representations, improving inference throughput.
Representative directions include activation-aware weight quantization (AWQ)~\cite{lin2023awq}, extreme compression with additive quantization~\cite{egiazarian2024extremecompressionlargelanguage}\cite{liu2025error},
and combining quantization with sparsity for further acceleration (GQSA)~\cite{zeng2025gqsagroupquantizationsparsity};
ABQ-LLM studies arbitrary-bit quantized inference to better trade accuracy for efficiency across hardware targets~\cite{zeng2025abqllmarbitrarybitquantizedinference}.
\textbf{Caching} targets dominant runtime/memory bottlenecks by reusing intermediate states.
For autoregressive LLMs, KV-cache quantization reduces cache memory and bandwidth~\cite{liu2024minicachekvcachecompression}.
For diffusion transformers (DiT), caching reuses redundant computation across timesteps, including layer caching~\cite{ma2024learningtocacheacceleratingdiffusiontransformer}\cite{liu2024foldgpt},
token-wise feature caching~\cite{zou2025acceleratingdiffusiontransformerstokenwise}, adaptive caching for video generation~\cite{kahatapitiya2024adaptivecachingfastervideo},
and error-aware cache correction with timestep adjustment~\cite{peng2025ertacacheerrorrectificationtimesteps}.

\section{Conclusion}
We present \textbf{S2O}, a fine-grained attention sparsification mechanism for long-context inference.
S2O enables non-contiguous, importance-driven token loading and an online skip strategy that safely prunes low-contribution block under a controlled error budget, moving beyond conventional fixed block-granularity designs.

\bibliographystyle{plainnat}
\bibliography{main}

\clearpage

\beginappendix

\section{LLM Usage}
Large Language Models were used solely to refine the manuscript’s language, including sentence rephrasing, grammar checking, and improving readability. The LLM was not involved in ideation, methodology, experimental design, or data analysis. All scientific content, concepts, and analyses were developed by the authors, who take full responsibility for the manuscript. The LLM’s role was limited to linguistic polishing, with strict adherence to ethical guidelines and avoidance of plagiarism or scientific misconduct.

\section{Permutation Overhead Analysis}
\label{sec:perm_overhead_analysis}

\subsection{Why Online Permutation Adds Negligible Overhead}
\label{sec:appendix_perm_overhead}

A natural concern of permutation-based sparsification is the potential overhead introduced by (i) explicit permutation, (ii) extra data movement, and (iii) irregular memory access that may reduce GPU efficiency. In this appendix, we explain why our online permutation incurs negligible overhead in practice.

\textbf{No physical permutation: permutation is implemented as index-driven loading.}
We do \textbf{not} physically permute $Q/K/V$ in memory, nor do we materialize permuted tensors. Instead, permutation is converted into a coordinate-driven loading policy: the attention kernel reads $Q/K/V$ from their original storage, but uses permuted indices for gather-style addressing and directly computes under the permuted schedule (Fig.~\ref{fig:online_load}). This eliminates the cost of explicit tensor permutation and large-scale memory copies.

\textbf{Metadata is lightweight and bandwidth-negligible.}
Online permutation only generates lightweight metadata (e.g., row indices for $Q$ and selected column indices for $KV$). Compared to the numeric $Q/K/V$ tensors, the index payload is tiny. For example, with $\texttt{head\_dim}=128$, a typical $K/V$ tile load can be as large as $2\times(128,128)$ (for $K$ and $V$), while the extra index read is only $(1,128)$ in scale, leading to negligible bandwidth overhead.

\textbf{Even with non-consecutive token indices, the underlying loads can remain regular and efficient.}
A GPU thread typically issues vectorized loads at a $128$-bit ($16$\,B) granularity; for BF16 ($2$\,B/elem), each load covers $8$ consecutive elements. Hence, for a token feature of dimension $D{=}64$, reading $64$ BF16 elements amounts to $8$ such vector loads, which can be viewed as $8$ threads collaboratively fetching one token (each responsible for $8$ contiguous elements). A warp contains $32$ threads and can therefore cover roughly $32/8{=}4$ tokens in parallel. Crucially, these tokens do not need to be adjacent in memory: as long as each token is stored contiguously and aligned along its feature dimension, the hardware still observes vectorized, coalesced transactions. Therefore, our offline token permutation does not require physical token contiguity; by preserving intra-token contiguity, it adheres to the vectorized/coalesced access principles and effectively utilizes memory bandwidth.

\textbf{The compute pipeline is unchanged: only iteration order and mapping change.}
Our kernel follows the FlashAttention execution pattern: for each $Q$ tile, we iterate over selected $K/V$ tiles, load them into SRAM/shared memory, and accumulate softmax statistics online. Permutation only changes the tile iteration order and the mapping from logical tile ids to physical offsets. The load--compute--accumulate pipeline and tiling configuration remain unchanged, introducing no extra synchronizations or additional passes.

\textbf{Empirical validation via ABCD decomposition.}
To precisely quantify the overhead of each component in our sparse attention pipeline, we design an ABCD decomposition benchmark. Starting from the dense FlashAttention kernel (Method A), each subsequent variant adds exactly one new mechanism, isolating its marginal cost:

\begin{itemize}
  \item \textbf{A (Dense):} Standard FlashAttention-2 tiled attention with contiguous memory access.
  \item \textbf{B (Indexed sequential):} Replaces contiguous tile loading with index-driven gather (\texttt{cp\_async} indexed load). The indices are the identity sequence $[0,1,2,\ldots]$, so the physical K/V access order is preserved; this isolates the overhead of adding the index-based addressing layer itself (the extra index-tensor load and the \texttt{cp\_async} indexed-mode instruction path), independent of any access-pattern irregularity.
  \item \textbf{C (Indexed sorted):} Identical to B except that the indices are pre-sorted by importance. Compared with B's identity order, C's indices induce a genuinely non-contiguous K/V access order, so the B$\to$C delta directly isolates the kernel-side cost of non-contiguous K/V access.
  \item \textbf{D (Early-stop logic):} Adds the deferred early-stop ratio check (computed after $P \cdot V$ GEMM, checked at the start of the next iteration) on top of C, measuring the overhead of the early-stopping control flow.
\end{itemize}

\begin{table}[t]
  \centering
  \begin{adjustbox}{max width=\linewidth}
  \begin{tabular}{l|rrrrr}
    \toprule
    Method & $N{=}8192$ & $N{=}16384$ & $N{=}32768$ & $N{=}65536$ & $N{=}131072$ \\
    \midrule
    A: Dense           &   3.38 &  11.95 &  44.67 & 175.40 & 717.49 \\
    B: Indexed seq     &   3.65 &  12.63 &  47.33 & 186.56 & 759.61 \\
    C: Indexed sorted  &   3.65 &  12.65 &  47.20 & 184.34 & 744.75 \\
    D: EarlyStop(off)  &   3.93 &  13.73 &  51.31 & 200.10 & 807.92 \\
    \bottomrule
  \end{tabular}
  \end{adjustbox}
  \caption{\textbf{ABCD decomposition (Causal).} Kernel-only latency (ms) on A100, bf16, $B{=}1$, $H{=}32$, $D{=}128$, seg\_len$=1024$. ``vs A'' = A kernel time / this kernel time.}
  \label{tab:abcde_causal}
\end{table}

\begin{table}[t]
  \centering
  \begin{adjustbox}{max width=\linewidth}
  \begin{tabular}{l|rrrrr}
    \toprule
    Method & $N{=}8192$ & $N{=}16384$ & $N{=}32768$ & $N{=}65536$ & $N{=}131072$ \\
    \midrule
    A: Dense           &   5.89 &  22.27 &  86.29 & 339.17 & 1344.97 \\
    B: Indexed seq     &   6.26 &  23.43 &  91.04 & 357.93 & 1421.07 \\
    C: Indexed sorted  &   6.24 &  23.53 &  91.02 & 358.14 & 1425.71 \\
    D: EarlyStop(off)  &   6.81 &  25.33 &  98.09 & 385.92 & 1565.86 \\
    \bottomrule
  \end{tabular}
  \end{adjustbox}
  \caption{\textbf{ABCD decomposition (Non-causal).} Same setup as Table~\ref{tab:abcde_causal}.}
  \label{tab:abcde_nocausal}
\end{table}

\begin{table}[t]
  \centering
  \begin{adjustbox}{max width=\linewidth}
  \begin{tabular}{l|cc|l}
    \toprule
    Transition & Causal & Non-causal & Description \\
    \midrule
    A $\to$ B & $\sim$5--8\% & $\sim$5--6\% & Indexed gather vs.\ contiguous tiled copy \\
    B $\to$ C & $\sim$0\% & $\sim$0\% & Sorted gather vs.\ sequential gather \\
    C $\to$ D & $\sim$8--9\% & $\sim$8--10\% & Deferred early-stop ratio check \\
    \midrule
    \textbf{A $\to$ D (total)} & \textbf{$\sim$13--16\%} & \textbf{$\sim$14--16\%} & \textbf{Full S2O machinery overhead} \\
    \bottomrule
  \end{tabular}
  \end{adjustbox}
  \caption{\textbf{Per-component marginal overhead} (kernel only, relative to the previous variant).}
  \label{tab:abcde_marginal}
\end{table}

\begin{figure}[t]
    \centering
    \includegraphics[width=0.48\linewidth]{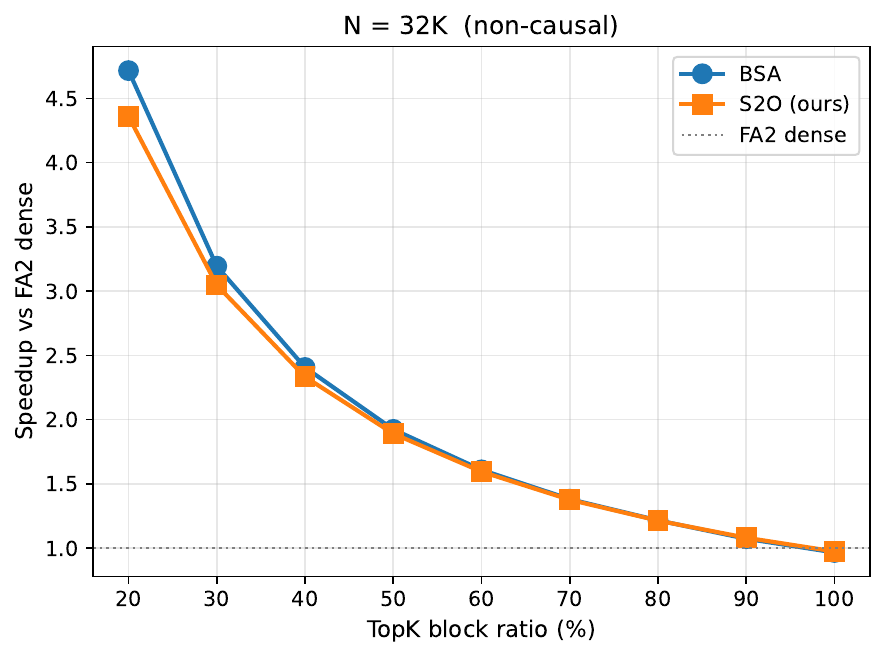}
    \hfill
    \includegraphics[width=0.48\linewidth]{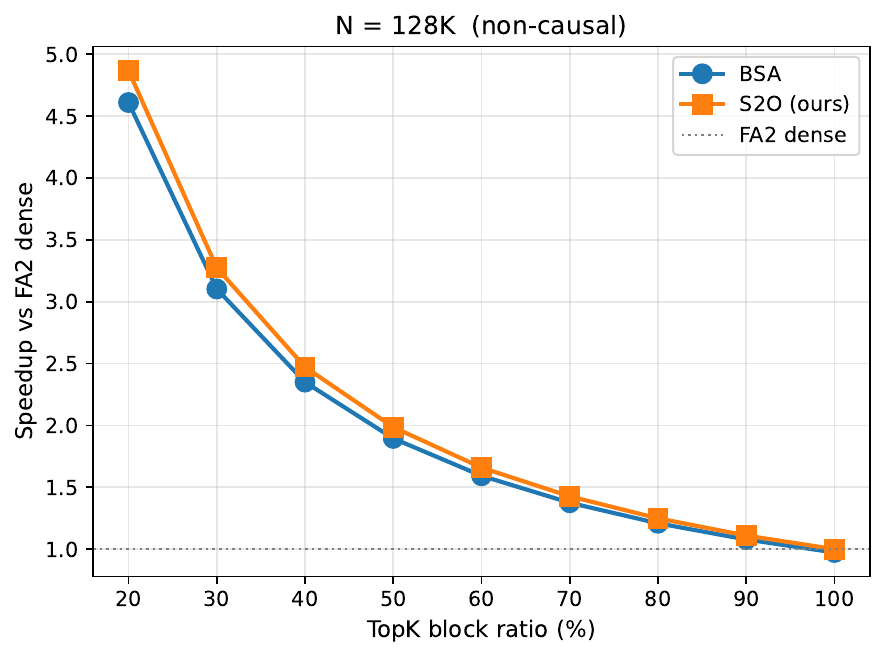}
    \caption{\textbf{Comparison with BSA under matched sparsity.}
    Speedup over FA2 dense under different TopK block ratios for non-causal attention.
    Left: sequence length $N=32$K. Right: sequence length $N=128$K.
    Under matched sparsity settings, S2O achieves runtime comparable to BSA, and the measured speedup remains largely aligned with the actual sparsity level. This suggests that sparsity beyond block granularity is practical and can be effectively translated into wall-clock acceleration.}\label{fig:bsa_compare_matched_sparsity}
\end{figure}

\textbf{Kernel provenance and controlled overhead.} Kernel A is built directly from the official FlashAttention-2 repository, while B, C, and D are implemented as minimal, strictly additive modifications on top of A, sharing the same tile sizes, warp/stage configuration, and online-softmax update. Therefore, the latency difference between any two adjacent variants can be attributed to exactly one newly introduced mechanism. All variants are run with \texttt{threshold}$=-1$ (i.e., no actual skipping), so they perform identical FLOPs as dense attention. Hence, any latency difference comes purely from the added mechanism itself. As shown in Tables~\ref{tab:abcde_causal}, \ref{tab:abcde_nocausal}, and \ref{tab:abcde_marginal}, the extra overhead introduced by S2O is overall small and well controlled: indexed gather adds only about $5$--$8\%$ overhead over dense attention, sorting is nearly negligible, and the deferred early-stop logic contributes about another $8$--$10\%$. Overall, even in the worst case without any actual skipping, the full S2O machinery increases kernel latency by only about $13$--$16\%$, indicating that the added mechanisms are lightweight and their implementation overhead is controllable in practice.

\textbf{Comparison with BSA under matched sparsity.}  
As shown in Fig.~\ref{fig:bsa_compare_matched_sparsity}, we compare S2O and BSA under matched sparsity settings at two sequence lengths, with $N=32$K on the left and $N=128$K on the right. Specifically, we adopt a top-$k$ selection strategy based on Kernel~C and tune both methods to operate at approximately the same effective sparsity, so that the comparison is not confounded by different amounts of computation removed. Across both sequence lengths, S2O achieves runtime comparable to BSA, while the measured speedup remains largely aligned with the actual sparsity level. These results suggest that going beyond block-level sparsity is feasible in practice: finer-grained sparsity does not introduce noticeable extra overhead, and can still be effectively translated into wall-clock acceleration. In the following, we further analyze in DiT the additional gains enabled by lower sparsity granularity.

\subsection{Why Online Permutation Integrates Seamlessly with FlashAttention}
\label{sec:appendix_perm_fa_compat_en}

\textbf{FlashAttention recap: tile streaming with an online softmax.}
FlashAttention accelerates attention by streaming over $K/V$ in tiles and keeping only a small working set on-chip. Concretely, for each query tile $Q_i$, the kernel iterates over a sequence of key/value tiles $\{(K_j,V_j)\}$, loads each tile into SRAM/shared memory, computes the partial score block $S_{ij}=Q_iK_j^\top$, and updates the output using an online softmax with running statistics (e.g., row-wise max and normalizer) before moving to the next tile. This load--compute--accumulate pipeline is highly structured: performance comes from (i) fixed-shape tiling, (ii) aligned/coalesced global-memory transactions per tile, and (iii) a single-pass streaming update without materializing the full $L\times L$ attention matrix.

\textbf{Where permutation traditionally hurts: physical permutation and extra passes.}
A straightforward way to apply permutation is to physically permute $Q/K/V$ into new contiguous layouts, or to scatter/gather them at very fine granularity. Such implementations introduce extra kernels, global-memory copies, and/or irregular accesses that break the tightly optimized FlashAttention dataflow. In practice, the overhead typically comes from \emph{changing the data layout} rather than changing the tile traversal order.

\textbf{Our key idea: replace physical permutation with coordinate scheduling.}
Our online permutation does not modify the FlashAttention compute pipeline. Instead, it replaces the \textbf{tile enumeration policy}:
\begin{itemize}
  \item \textbf{Baseline FlashAttention:} visit tiles in a canonical order (e.g., increasing $j$) with contiguous offsets.
  \item \textbf{Online permutation:} visit tiles in a permuted schedule determined by lightweight index arrays (e.g., $Q_{\mathrm{perm}}$, $K_{\mathrm{perm}}$), and map each logical tile id to its physical offset via index-addressed gathers (Fig.~\ref{fig:online_load}).
\end{itemize}
Crucially, within each visited tile, we still load contiguous fragments of $Q/K/V$ with the same alignment as FlashAttention. Thus, the GPU observes regular tile-shaped memory transactions, while the logical computation order is globally permuted.

\textbf{What is changed vs.\ unchanged.}
Online permutation changes only: (i) the iteration order of $(K_j,V_j)$ tiles for a given $Q_i$ (and optionally the ordering of $Q_i$ tiles), and (ii) the address mapping from a logical tile id to a physical base pointer. Everything else remains identical to FlashAttention: the tile sizes, the number of stages/warps, the single-pass online softmax update, and the synchronization structure. As a result, online permutation can be integrated as a thin addressing layer on top of a FlashAttention-style kernel, preserving its efficiency while enabling global, index-guided prioritization of high-contribution tiles.

\section{Additional Experimental Details}
\label{sec:append}

\subsection{Baseline hyperparameters}
\label{sec:append_hyper}

\noindent\textbf{Unified settings.}
To ensure fair comparisons, all methods are evaluated with the same model weights, maximum context length, batch size, dtype, attention backend, and decoding configuration. Whenever applicable, we use the same input samples and random seeds across methods. When reporting sparsity and approximation error across different context lengths, we use the first example in LongBench v2 and construct each length setting via truncation.

\noindent\textbf{PBS-Attn.}
We follow the default PBS-Attn configuration: block size $B=128$, segment size $S=256$, and a block-selection threshold of $0.9$ in all experiments. We also keep the default block grouping/selection procedure as in the original implementation.

\noindent\textbf{FlexPrefill.}
We set the sparse-pattern threshold $\tau=0.1$ for all models. We use the head-wise budget controller $\gamma$ to adapt the compute budget online, and set $\gamma=0.9$ by default.

\noindent\textbf{X-Attn.}
We use the default X-Attn configuration with threshold $\tau=0.9$ and stride $16$ for all models.

\subsection{Additional LLM Experiments}
\label{sec:append_llm_extra}

\begin{table}[t]
  \centering
  \scriptsize
  \renewcommand{\arraystretch}{1.10}
  \setlength{\tabcolsep}{4.2pt}
  \begin{adjustbox}{width=\linewidth}
  \begin{tabular}{lcccccc}
    \toprule
    Methods & 8K & 16K & 32K & 64K & 128K & Avg. \\
    \midrule

    \rowcolor{modelgray}
    \multicolumn{7}{c}{\textbf{Llama-3.1-8B}} \\
    \midrule
    Full &
    \spcell{91.53}{0.000} &
    \spcell{87.61}{0.000} &
    \spcell{89.30}{0.000} &
    \spcell{87.96}{0.000} &
    \spcell{75.42}{0.000} &
    \spcell{86.36}{0.000} \\

    FlexPrefill &
    \spcell{86.89}{0.684} &
    \spcell{86.51}{0.729} &
    \spcell{81.86}{0.790} &
    \spcell{76.35}{0.828} &
    \spcell{72.28}{0.840} &
    \spcell{80.78}{0.774} \\

    X-Attn &
    \spcell{91.10}{0.456} &
    \spcell{88.62}{0.559} &
    \spcell{87.90}{0.652} &
    \spcell{84.67}{0.653} &
    \spcell{\textbf{73.15}}{0.737} &
    \spcell{85.09}{0.611} \\

    PBS &
    \spcell{91.06}{0.356} &
    \spcell{88.46}{0.468} &
    \spcell{85.69}{0.560} &
    \spcell{78.36}{0.668} &
    \spcell{67.01}{0.747} &
    \spcell{82.12}{0.560} \\

    \rowcolor{oursblue}
    \textbf{Ours} &
    \spcell{\textbf{93.03}}{0.324} &
    \spcell{\textbf{90.33}}{0.591} &
    \spcell{\textbf{87.91}}{0.768} &
    \spcell{\textbf{85.29}}{0.875} &
    \spcell{72.40}{0.932} &
    \spcell{\textbf{85.79}}{0.698} \\

    \midrule
    \rowcolor{modelgray}
    \multicolumn{7}{c}{\textbf{Qwen3-8B}} \\
    \midrule
    Full &
    \spcell{85.77}{0.000} &
    \spcell{81.81}{0.000} &
    \spcell{81.40}{0.000} &
    \spcell{73.66}{0.000} &
    \spcell{69.50}{0.000} &
    \spcell{78.43}{0.000} \\

    FlexPrefill &
    \spcell{71.65}{0.675} &
    \spcell{73.89}{0.704} &
    \spcell{75.38}{0.752} &
    \spcell{72.65}{0.777} &
    \spcell{68.51}{0.826} &
    \spcell{72.42}{0.747} \\

    X-Attn &
    \spcell{85.63}{0.458} &
    \spcell{82.25}{0.534} &
    \spcell{\textbf{81.60}}{0.609} &
    \spcell{73.18}{0.668} &
    \spcell{69.91}{0.737} &
    \spcell{78.51}{0.601} \\

    PBS &
    \spcell{85.56}{0.413} &
    \spcell{79.34}{0.504} &
    \spcell{80.95}{0.579} &
    \spcell{70.70}{0.653} &
    \spcell{67.90}{0.733} &
    \spcell{76.89}{0.576} \\

    \rowcolor{oursblue}
    \textbf{Ours} &
    \spcell{\textbf{85.80}}{0.191} &
    \spcell{\textbf{82.73}}{0.440} &
    \spcell{80.34}{0.644} &
    \spcell{\textbf{73.95}}{0.796} &
    \spcell{\textbf{69.97}}{0.893} &
    \spcell{\textbf{78.56}}{0.593} \\

    \bottomrule
  \end{tabular}
  \end{adjustbox}

  \caption{\textsc{RULER} accuracy (top) and prefill-stage sparsity ratio (bottom) across input lengths (8K--128K). Accuracy is averaged over tasks for each length; sparsity ratios are computed under identical decoding settings.}
  \label{tab:ruler_acc_sparsity}
\end{table}

\begin{table}
  \centering
  \renewcommand{\arraystretch}{1.05}
  \begin{adjustbox}{width=\linewidth}
  \begin{tabular}{lcc}
    \toprule
    Method & Llama-3.1-8B & Qwen3-8B \\
    \midrule
    Full        & 30.22 & 34.19 \\
    FlexPrefill & 26.44 & 30.21 \\
    X-Attn      & 29.82 & 32.00 \\
    PBS         & 30.41 & 32.00 \\
    \rowcolor{oursblue}
    Ours        & \textbf{31.41} & \textbf{32.60} \\
    \bottomrule
  \end{tabular}
  \end{adjustbox}
  \caption{\textsc{LongBench v2} average score.}
  \label{tab:longbenchv2}
\end{table}

\paragraph{RULER.}
Under the same settings as Table~\ref{tab:ruler_acc_sparsity}, we report the prefill-stage sparsity ratio and accuracy. RULER spans a wide range of context lengths, enabling a more comprehensive evaluation of how sparsification strategies behave and pay off in typical-length regimes. Our results show that even at shorter sequence lengths, our method still achieves high sparsity and consistently translates it into end-to-end speedups, indicating that the proposed coordinate scheduling and early-stopping mechanism remains effective beyond the ultra-long-context setting and provides a favorable benefit--overhead trade-off in practice.

\paragraph{LongBench v2.}
On LongBench v2 (Table~\ref{tab:longbenchv2}), our method continues to deliver strong accuracy while leveraging high prefill sparsity for acceleration, demonstrating robust generalization to a broader set of long-context tasks.

\subsection{Single-pass Variant}
\label{sec:append_fused_pass12}
For completeness, we provide a variant that does not reorder $Q$ and fuses
Pass-1 (dense intra-segment causal initialization) and Pass-2 (coordinate-scheduled prefix traversal with early stopping)
into a single streaming procedure.
Unlike the two-pass implementation that materializes intermediate online-softmax states $(A, Lbuf, Mbuf)$ in (H)BM,
the fused variant keeps the per-query states $(m,\ell,acc)$ in registers and updates them sequentially:
(i) it first scans intra-segment causal tiles to initialize the states (equivalent to Pass-1),
(ii) then computes a small dense local window for boundary stability,
and (iii) finally traverses historical prefix tiles in the retrieved order with a monotone-gain early-stopping rule.
See pseudocode in Alg.~\ref{alg:fused_pass12_noqreorder}.

\providecommand{\spcell}[2]{\makecell{#1\\{\scriptsize\textcolor{black!45}{#2}}}}
\providecommand{\spcellna}[1]{\makecell{#1\\{\scriptsize\textcolor{black!45}{--}}}}

\definecolor{algteal}{RGB}{0,128,128}
\algrenewcommand\algorithmiccomment[1]{\hfill\textcolor{algteal}{\footnotesize\itshape \#~#1}}

\begin{algorithm}[t]
\caption{Fused Pass-1+Pass-2 (Single-pass, No $Q$ Reordering): Coordinate-Scheduled Sparse Attention}
\label{alg:fused_pass12_noqreorder}
\footnotesize
\begin{algorithmic}[1]
\AlgInput{$Q,K,V$; $KV\algsub{perm}$ (Alg.~\ref{alg:ordering}); segment length $S$; local window $W\le S$; threshold $\tau$; tile sizes $(B_M,B_N)$}
\AlgOutput{$O$ (causal)}

\For{$n \gets 0$ \textbf{to} $N-1$}
  \For{$h \gets 0$ \textbf{to} $H-1$}
    \Comment{scan queries in the original order, in $B_M$ tiles}
    \ForAll{\textbf{each} query tile $b$ in segment $n$}
      \State $q \gets Q(n,h,b)$
      \State $s \gets (-\infty,\,0,\,0)$

      \Comment{(Pass-1) dense intra-segment causal init}
      \ForAll{\textbf{each} key tile $t$ in segment $n$}
        \State $k \gets K(n,h,t)$;\quad $v \gets V(n,h,t)$
        \State $\mathcal{M}\gets \MaskCausal(q\text{ tile }b,\ k\text{ tile }t)$
        \State $s \gets \FlashOS\big(q,\ k,\ v;\ \mathcal{M},\ s\big)$
      \EndFor

      \Comment{(Pass-2) prefix traversal in ranked order with early stopping}
      \For{$T \in KV\algsub{perm}[n]$}
        \State $k_T \gets K[T]$;\quad $v_T \gets V[T]$
        \State $s' \gets \FlashOS\big(q,\ k_T,\ v_T;\ s\big)$
        \State $\Delta\ell \gets s'.\ell - s.\ell$
        \If{$\Delta\ell < \tau\cdot s.\ell$}
          \State \textbf{break}
        \EndIf
        \State $s \gets s'$
      \EndFor

      \State $O(n,h,b)\gets s.acc/s.\ell$
    \EndFor
  \EndFor
\EndFor

\Statex \textbf{return} $O$
\end{algorithmic}
\end{algorithm}

\subsection{Additional details on end-to-end operator performance}
\label{sec:e2e_operator_details}

\paragraph{Retrieval overhead vs.\ sequence length.}
Our operator consists of (i) a lightweight retrieval stage that computes segment-wise ranking indices (e.g., sorted key offsets and an optional query permutation), followed by (ii) the sparse attention computation. In our implementation, the dominant retrieval cost comes from scoring keys using per-segment query summaries, which scales approximately as $\mathcal{O}(L^2/S)$ where $L$ is the sequence length and $S$ is the segment length. The remaining parts (e.g., local query scoring and intra-segment sorting) contribute lower-order terms close to $\mathcal{O}(L)$ (and a small $\mathcal{O}(L\log S)$ sorting cost). Therefore, increasing $S$ reduces retrieval overhead roughly inversely, while increasing $L$ amplifies it superlinearly. In practice, we use a smaller segment length for short sequences to increase sparsity, and switch to a larger segment length as $L$ grows to reduce retrieval time, since at ultra-long contexts the sparsity ratio is often already very high and the retrieval overhead can become non-negligible relative to the sparse attention computation itself. Table~\ref{tab:retrieval_overhead_seglen_seqlen} reports the measured retrieval latency (ms) across $(L,S)$.

\begin{table}[t]
  \centering
  \setlength{\tabcolsep}{4pt}
  \renewcommand{\arraystretch}{1.05}
  \resizebox{\linewidth}{!}{%
  \begin{tabular}{rcccc c}
    \toprule
    \multirow{2}{*}{$L$} &
    \multicolumn{4}{c}{Retrieval (ms)} &
    \multirow{2}{*}{Flash latency (ms)} \\
    \cmidrule(lr){2-5}
      & $S{=}512$ & $S{=}1024$ & $S{=}2048$ & $S{=}4096$ & \\
    \midrule
    4K   & 0.254 & 0.255 & 0.256 & 0.246 & 0.756 \\
    8K   & 0.374 & 0.311 & 0.266 & 0.315 & 1.720 \\
    16K  & 0.785 & 0.545 & 0.386 & 0.328 & 5.712 \\
    32K  & 2.415 & 1.410 & 0.920 & 0.636 & 21.09 \\
    64K  & 8.644 & 4.646 & 2.667 & 1.713 & 81.48 \\
    128K & 33.238 & 16.906 & 9.017 & 5.122 & 320.2 \\
    \bottomrule
  \end{tabular}%
  }
  \caption{Measured retrieval latency (ms) as a function of segment length $S$ and sequence length $L$, with Flash latency shown on the right.}
  \label{tab:retrieval_overhead_seglen_seqlen}
\end{table}

\section{Additional Attention Heatmaps}
\label{app:heatmaps}

\begin{figure*}[t]
  \centering
  \setlength{\tabcolsep}{1pt}
  \renewcommand{\arraystretch}{1.0}
  \footnotesize
  \def\imgw{0.24\textwidth}

  \begin{tabular}{@{}c@{\hspace{2pt}}c@{\hspace{2pt}}c@{\hspace{2pt}}c@{}}
  \includegraphics[width=\imgw]{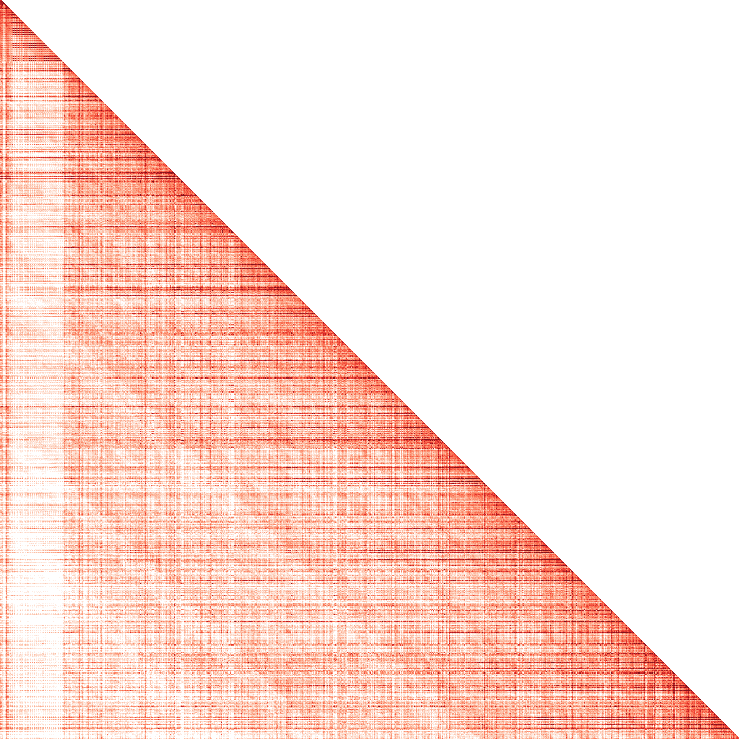}\phantomsubcaption\label{fig:qwen_grid_attn_a1} &
  \includegraphics[width=\imgw]{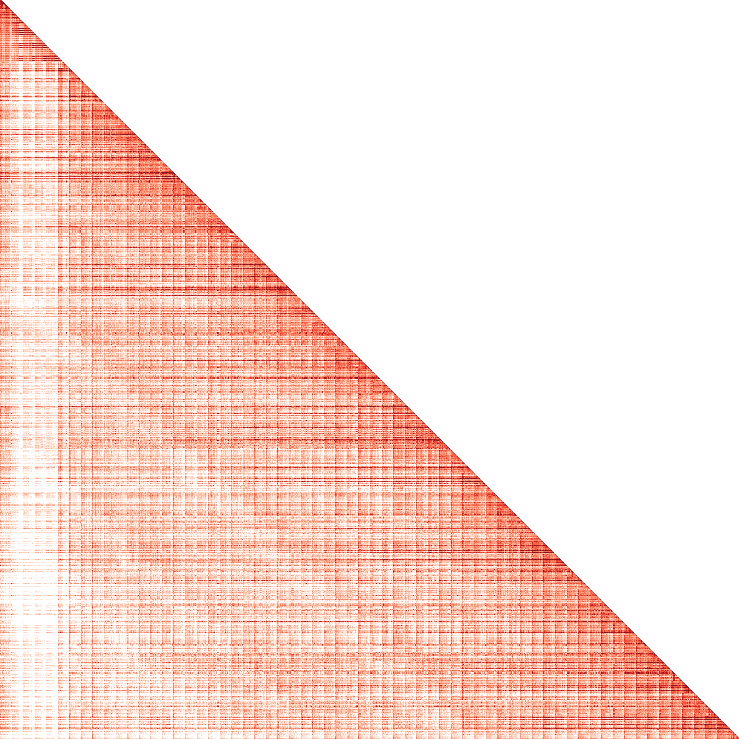}\phantomsubcaption\label{fig:qwen_grid_attn_b1} &
  \includegraphics[width=\imgw]{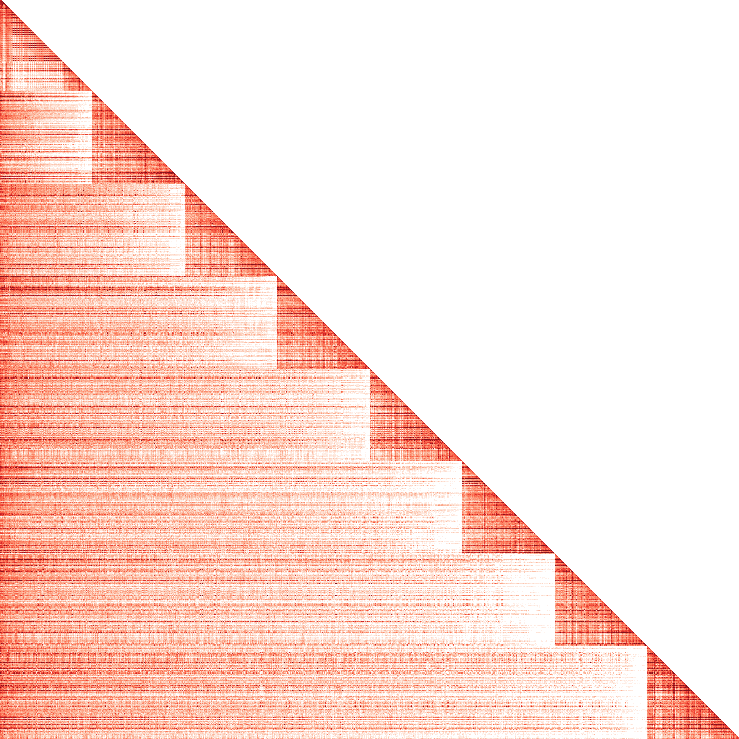}\phantomsubcaption\label{fig:qwen_grid_attn_c1} &
  \includegraphics[width=\imgw]{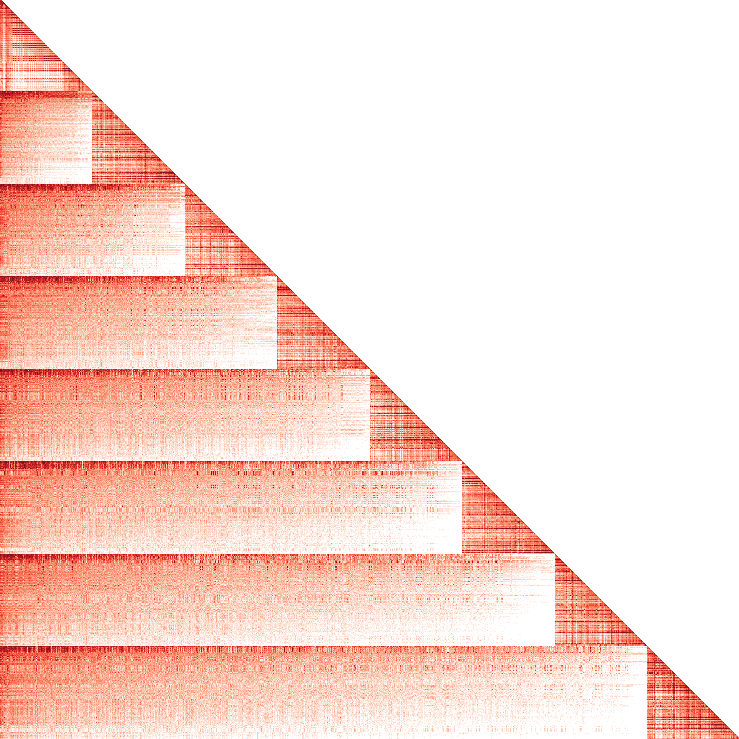}\phantomsubcaption\label{fig:qwen_grid_attn_d1} \\
  \includegraphics[width=\imgw]{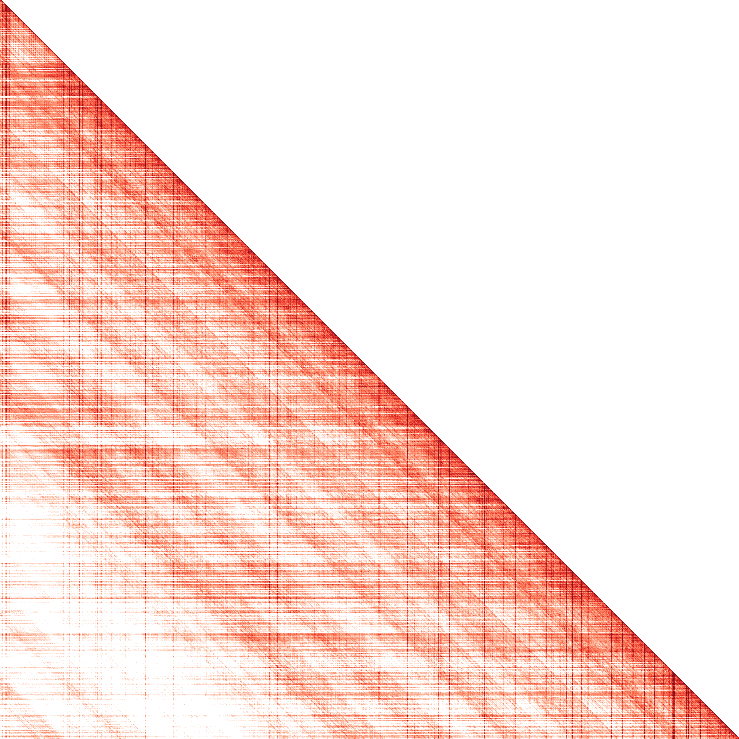}\phantomsubcaption\label{fig:qwen_grid_attn_a2} &
  \includegraphics[width=\imgw]{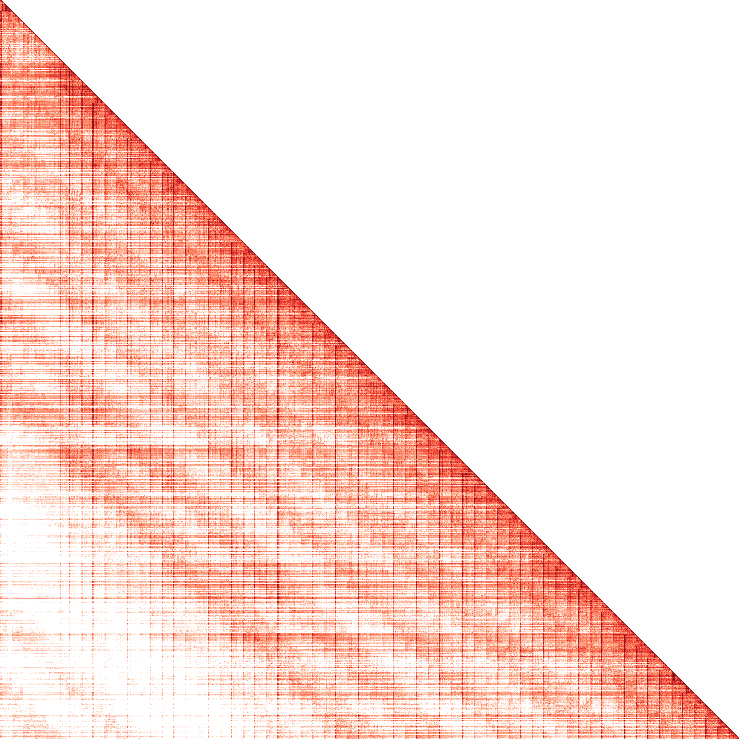}\phantomsubcaption\label{fig:qwen_grid_attn_b2} &
  \includegraphics[width=\imgw]{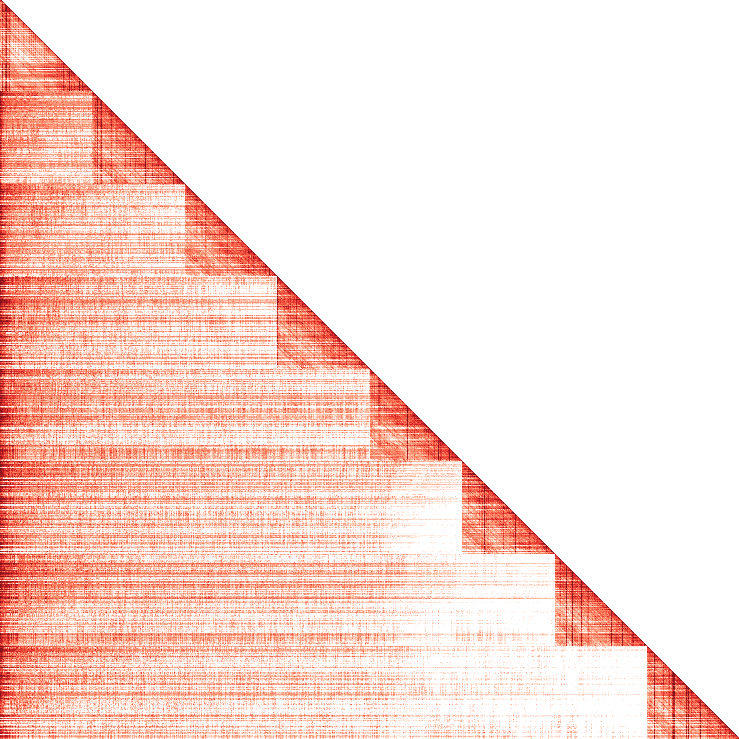}\phantomsubcaption\label{fig:qwen_grid_attn_c2} &
  \includegraphics[width=\imgw]{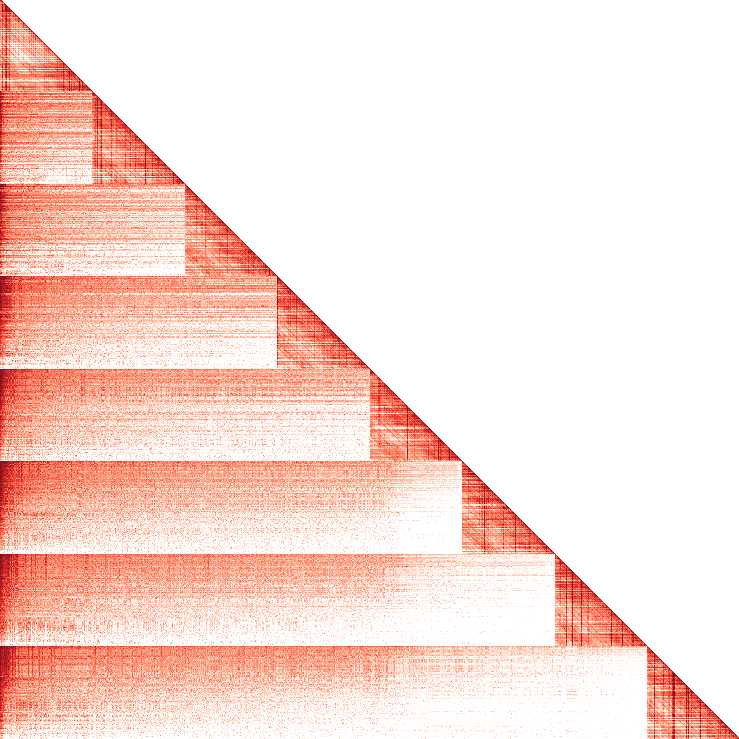}\phantomsubcaption\label{fig:qwen_grid_attn_d2} \\
  \textbf{(a) Original} &
  \textbf{(b) PBS} &
  \textbf{(c) Ours (w/o $Q$ reorder)} &
  \textbf{(d) Ours} \\
  \end{tabular}

  \caption[Additional Qwen attention heatmaps]{\textbf{Additional attention heatmaps on Qwen3-8B under different reordering strategies.}}
  \label{fig:qwen_grid_attn}
\end{figure*}

\begin{figure*}[t]
  \centering
  \setlength{\tabcolsep}{1pt}
  \renewcommand{\arraystretch}{1.0}
  \footnotesize
  \def\imgw{0.24\textwidth}

  \begin{tabular}{@{}c@{\hspace{2pt}}c@{\hspace{2pt}}c@{\hspace{2pt}}c@{}}
  \includegraphics[width=\imgw]{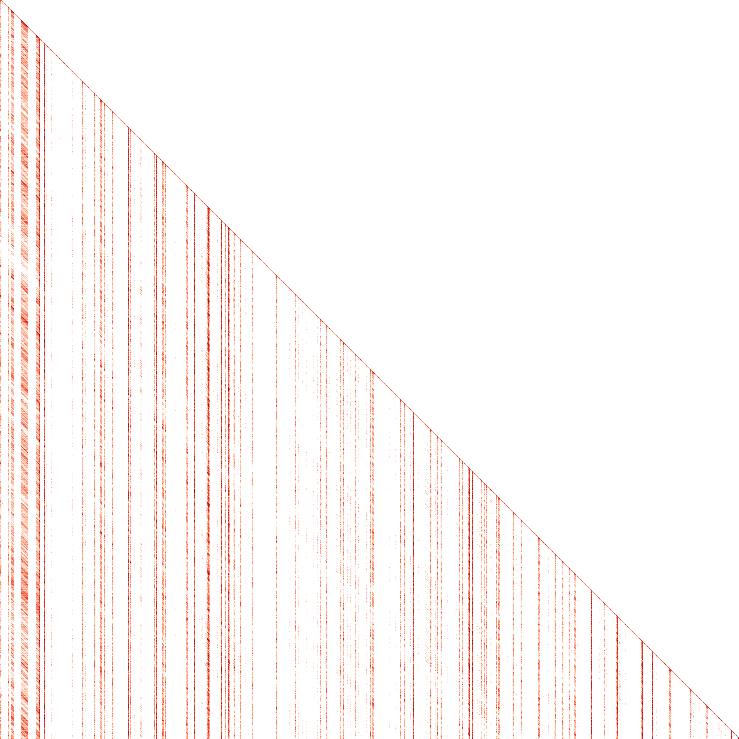}\phantomsubcaption\label{fig:llama_grid_attn_a1} &
  \includegraphics[width=\imgw]{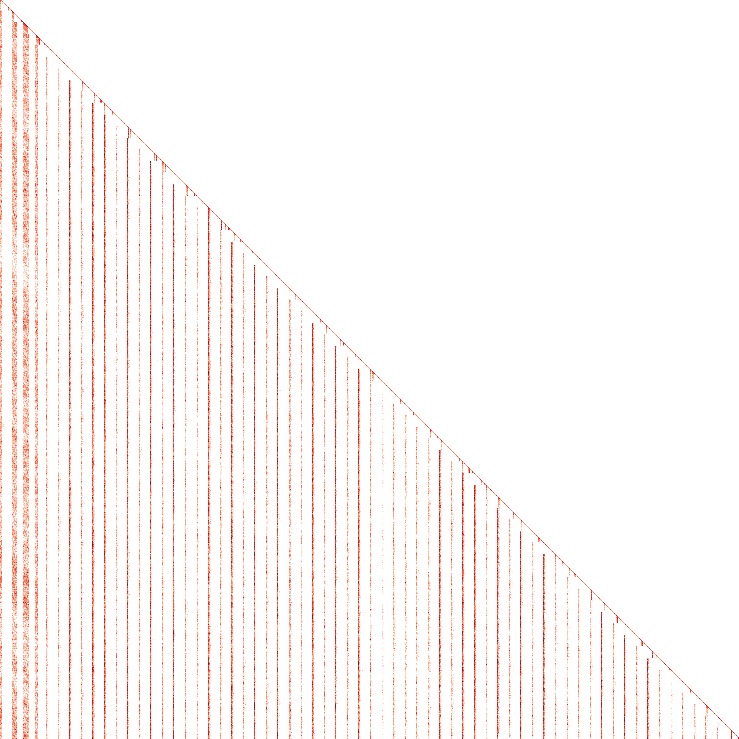}\phantomsubcaption\label{fig:llama_grid_attn_b1} &
  \includegraphics[width=\imgw]{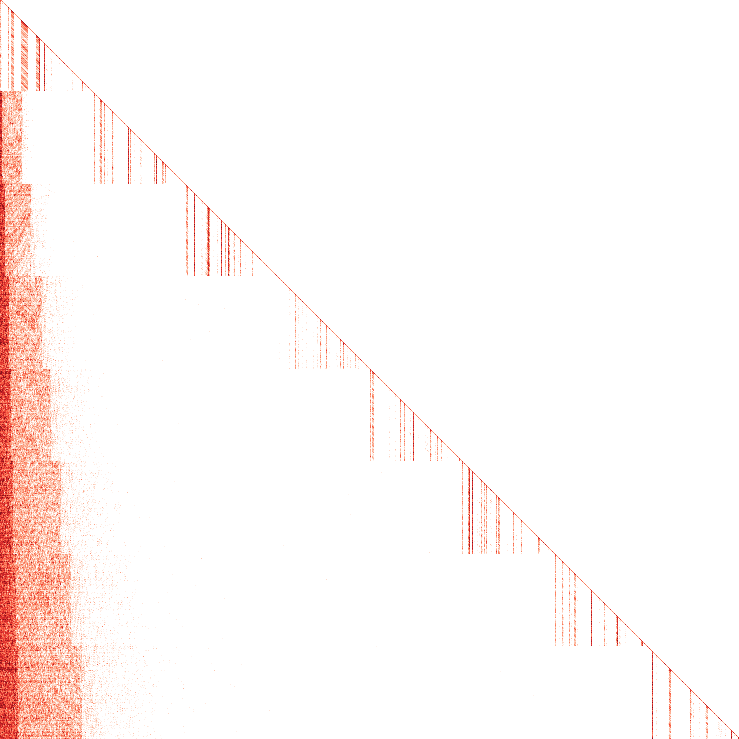}\phantomsubcaption\label{fig:llama_grid_attn_c1} &
  \includegraphics[width=\imgw]{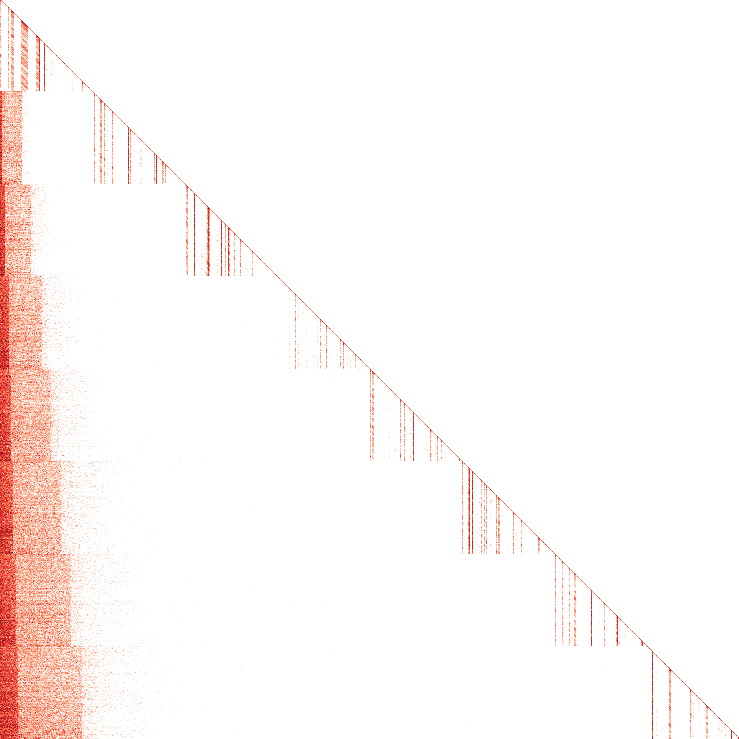}\phantomsubcaption\label{fig:llama_grid_attn_d1} \\
  \includegraphics[width=\imgw]{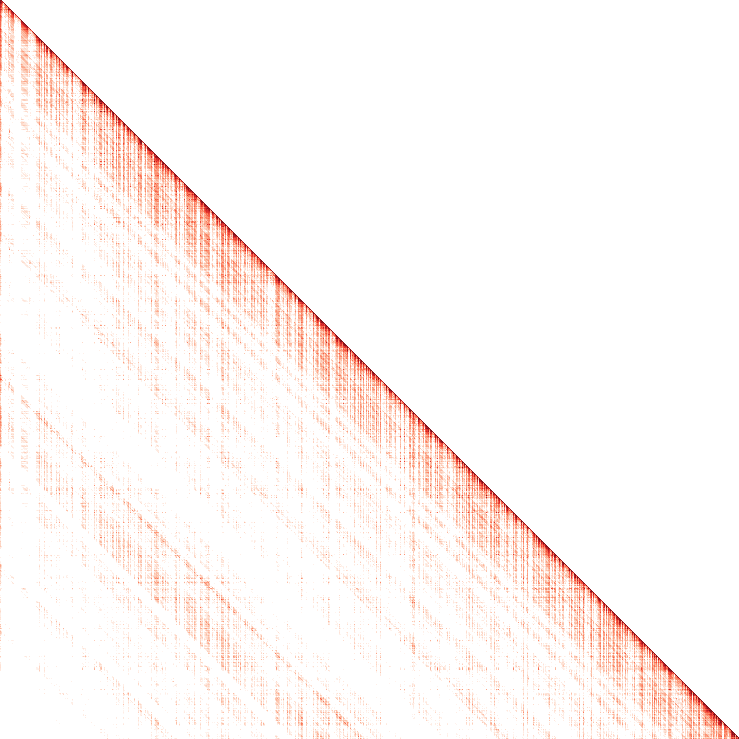}\phantomsubcaption\label{fig:llama_grid_attn_a2} &
  \includegraphics[width=\imgw]{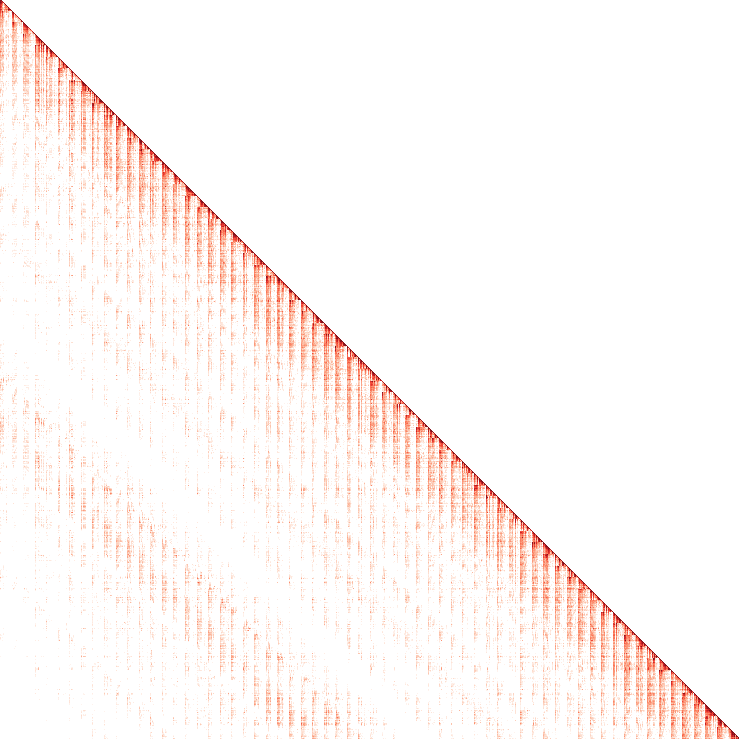}\phantomsubcaption\label{fig:llama_grid_attn_b2} &
  \includegraphics[width=\imgw]{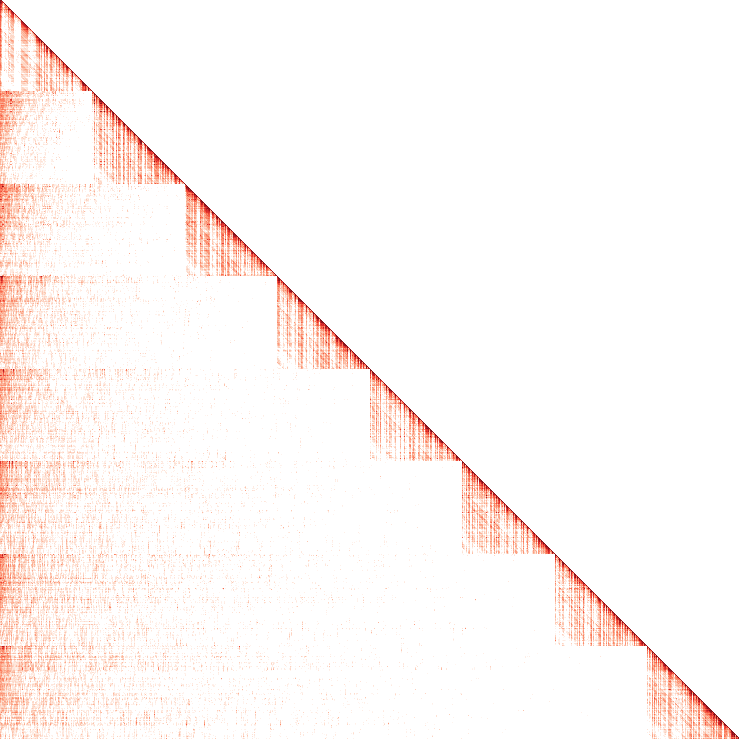}\phantomsubcaption\label{fig:llama_grid_attn_c2} &
  \includegraphics[width=\imgw]{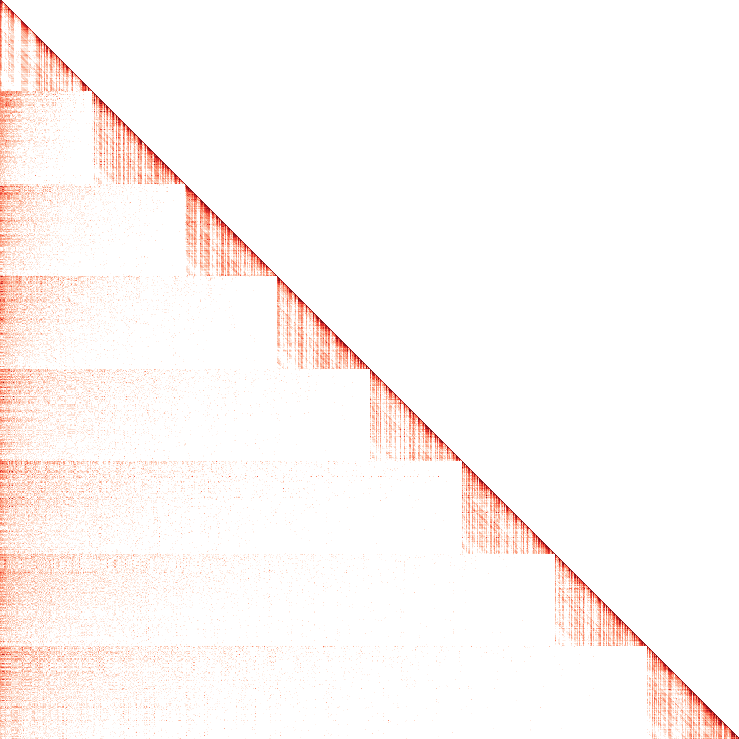}\phantomsubcaption\label{fig:llama_grid_attn_d2} \\
  \includegraphics[width=\imgw]{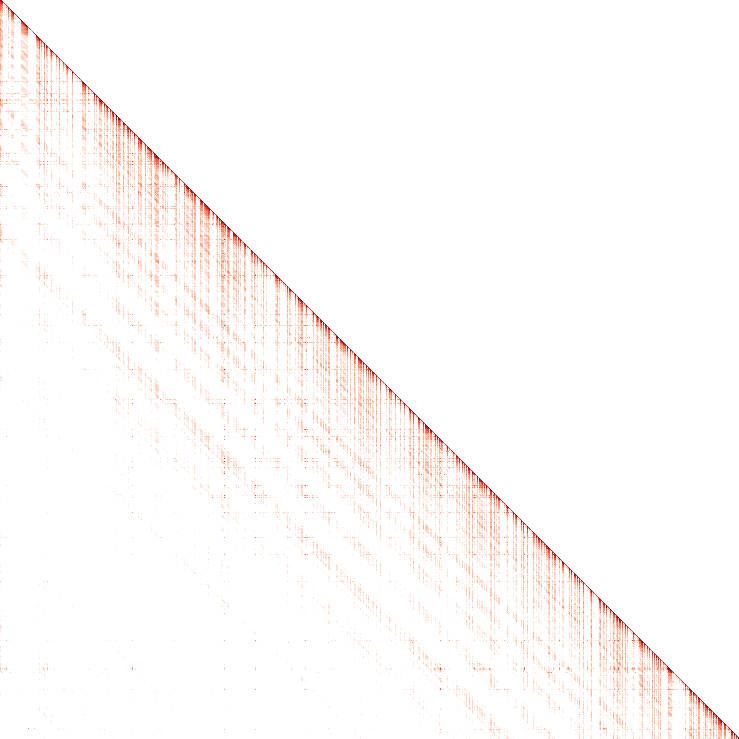}\phantomsubcaption\label{fig:llama_grid_attn_a3} &
  \includegraphics[width=\imgw]{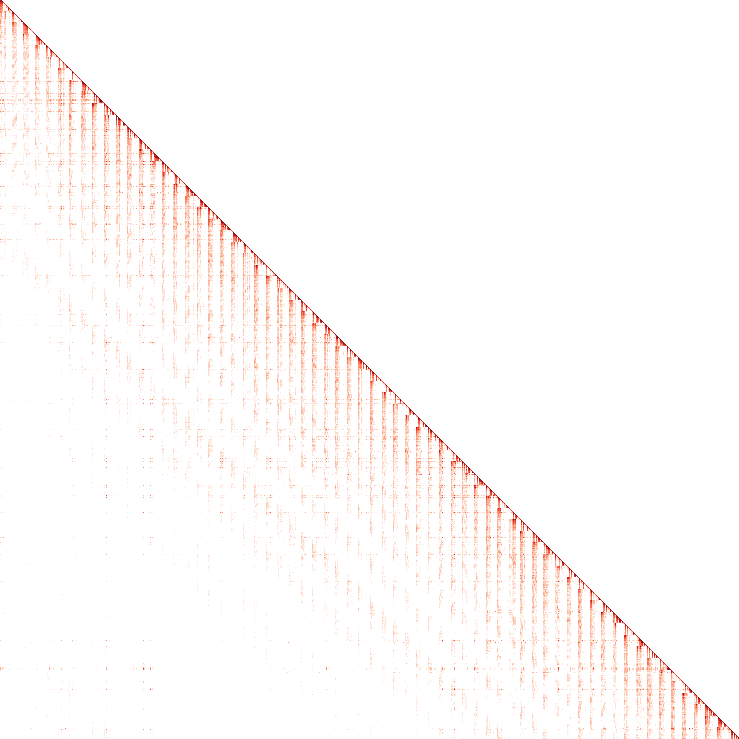}\phantomsubcaption\label{fig:llama_grid_attn_b3} &
  \includegraphics[width=\imgw]{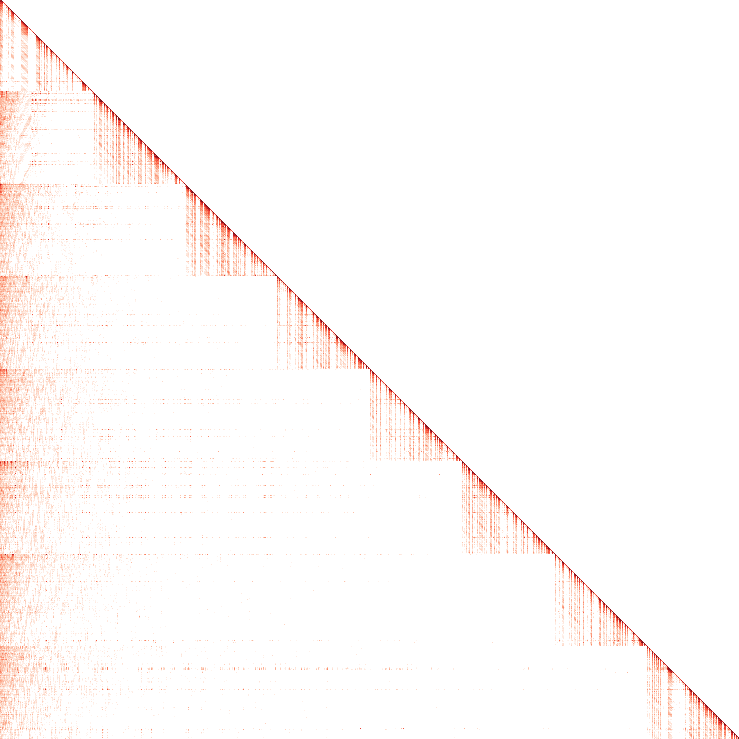}\phantomsubcaption\label{fig:llama_grid_attn_c3} &
  \includegraphics[width=\imgw]{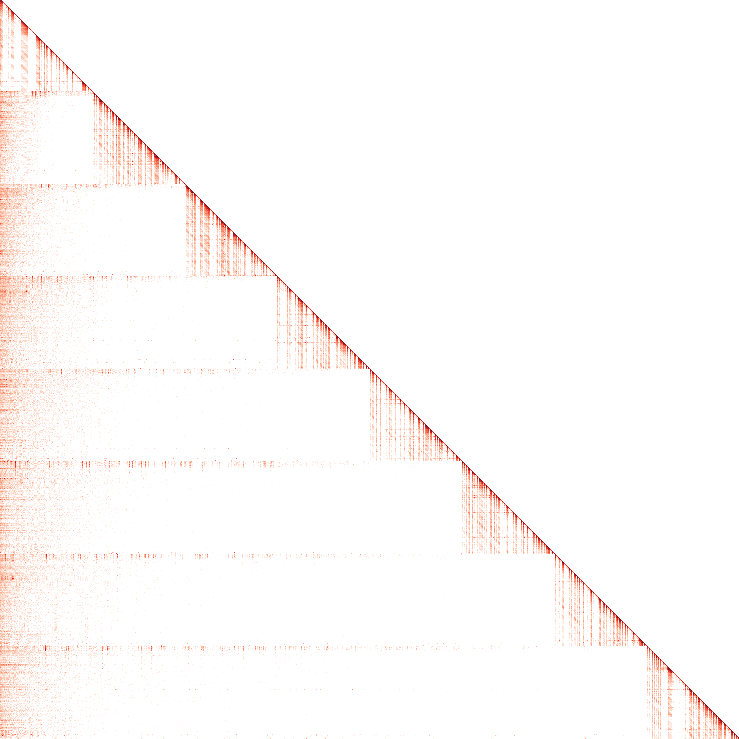}\phantomsubcaption\label{fig:llama_grid_attn_d3} \\[-2pt]
  \textbf{(a) Original} &
  \textbf{(b) PBS} &
  \textbf{(c) Ours (w/o $Q$ reorder)} &
  \textbf{(d) Ours} \\
  \end{tabular}

  \caption[Additional LLaMA attention heatmaps]{\textbf{Additional attention heatmaps on LLaMA-3.1-8B under different reordering strategies.}}
  \label{fig:llama_grid_attn}
\end{figure*}

We provide additional qualitative attention heatmaps for two representative backbones: \textbf{LLaMA-3.1-8B} and \textbf{Qwen3-8B}.
For each model, we compare four strategies: \textbf{Original}, \textbf{PBS} (local $K/V$ permutation), \textbf{Ours (w/o $Q$ permutation)} (global $K/V$ permutation only), and \textbf{Ours} (global permutation of both $Q$ and $K/V$), as shown in Fig.~\ref{fig:qwen_grid_attn} and Fig.~\ref{fig:llama_grid_attn}, respectively.
These heatmaps indicate that across a broader set of layers/heads and inputs, our method consistently compacts salient stripe-like structures toward the upper-left region, demonstrating strong usability and consistency.

\end{document}